\documentclass[11pt]{article}

\usepackage{iftex}
\ifPDFTeX
  \usepackage[T1]{fontenc}
  \usepackage[utf8]{inputenc}
\else
\fi
\usepackage{lmodern}

\usepackage[letterpaper,margin=1in]{geometry}
\usepackage{microtype}
\usepackage{fancyhdr}
\usepackage{titlesec}

\usepackage{amsmath,amssymb,amsthm,mathtools,bm}

\usepackage{graphicx}
\usepackage{booktabs}
\usepackage{multirow}
\usepackage{array}
\usepackage{caption}
\usepackage{subcaption}
\usepackage{xcolor}
\usepackage{tikz}
\usepackage{pgfplots}
\pgfplotsset{compat=1.18}
\usetikzlibrary{arrows.meta,positioning,fit,backgrounds,calc,shapes.geometric}

\usepackage{algorithm}
\usepackage{algpseudocode}

\usepackage{enumitem}
\usepackage{listings}
\usepackage{xspace}

\usepackage[numbers,sort&compress]{natbib}

\usepackage{url}
\usepackage[colorlinks=true,linkcolor=RoyalBlue,citecolor=RoyalBlue,urlcolor=RoyalBlue]{hyperref}
\usepackage[capitalize]{cleveref}
\usepackage{orcidlink}

\definecolor{RoyalBlue}{RGB}{30,80,160}
\definecolor{trgreen}{RGB}{20,120,70}
\definecolor{trred}{RGB}{170,40,40}
\definecolor{trgray}{RGB}{110,110,110}
\definecolor{axisA}{RGB}{31,119,180}   
\definecolor{axisE}{RGB}{214,39,40}    
\definecolor{axisB}{RGB}{44,160,44}    

\theoremstyle{definition}
\newtheorem{principle}{Design Principle}
\newtheorem{observation}{Observation}

\newcommand{\method}{\textsc{TriRoute}\xspace}
\newcommand{\router}{\ensuremath{g_\phi}}
\newcommand{\Real}{\mathbb{R}}

\newcommand{\softmax}{\operatorname{softmax}}
\newcommand{\sg}{\operatorname{sg}}                  
\newcommand{\gumbel}{\operatorname{Gumbel}}
\newcommand{\dattn}{a}       
\newcommand{\dexp}{e}        
\newcommand{\dbit}{b}        
\newcommand{\Ax}{\mathcal{A}}
\newcommand{\Ex}{\mathcal{E}}
\newcommand{\Bx}{\mathcal{B}}
\newcommand{\LM}{\mathcal{L}_{\mathrm{LM}}}
\newcommand{\Lbal}{\mathcal{L}_{\mathrm{bal}}}
\newcommand{\Lz}{\mathcal{L}_{z}}
\newcommand{\Lent}{\mathcal{L}_{\mathrm{ent}}}

\newcommand{\Cstar}{{C^{\star}}}

\lstdefinestyle{py}{
  language=Python,
  basicstyle=\ttfamily\footnotesize,
  keywordstyle=\color{RoyalBlue}\bfseries,
  commentstyle=\color{trgreen}\itshape,
  stringstyle=\color{trred},
  numbers=left,
  numberstyle=\tiny\color{trgray},
  numbersep=6pt,
  showstringspaces=false,
  breaklines=true,
  frame=single,
  rulecolor=\color{trgray!50},
  columns=fullflexible,
  keepspaces=true,
}

\title{\vspace{-1.2em}\bfseries \method: Unified Learned Routing for Joint\\[2pt]
Adaptive Attention, Experts, and KV-Cache Allocation}

\author{%
  Andrii Balashov\,\orcidlink{0009-0007-5833-0888}
  \quad
  Olena Ponomarova\,\orcidlink{0000-0003-1254-4403}\\[0.4em]
  \small
  Ukrainian State University of Science and Technologies\\[0.35em]
  \href{mailto:andbalashov@hotmail.com}{\texttt{andbalashov@hotmail.com}}
  \quad
  \href{mailto:o.a.ponomarova@ust.edu.ua}{\texttt{o.a.ponomarova@ust.edu.ua}}
}
\date{}

\begin{document}
\maketitle

\begin{abstract}
Conditional computation promises to decouple a language model's quality from its
per-token inference cost, but the leading techniques each act on a \emph{single}
axis and are developed in isolation: Mixture-of-Experts (MoE) sparsifies the
feed-forward network, Mixture-of-Depths (MoD) skips whole transformer blocks, and
KV-cache quantization compresses attention memory. We argue that these three
decisions---\emph{how much attention resolution}, \emph{which experts}, and
\emph{how many bits of cache} a token deserves---are strongly coupled and should
be made jointly. We introduce \method, a single lightweight controller that, for
every token at every layer, emits a coordinated policy over all three axes:
(i) an attention mode (skip / local / full), (ii) a sparse set of FFN experts
(including a null expert that recovers MoD), and (iii) a KV-cache bit-width that
governs how the token is remembered by future queries. The controller is trained
end-to-end with the language-modeling objective using a heterogeneous relaxation
scheme---Gumbel-Softmax with straight-through estimation for the categorical
attention and bit decisions, and load-balanced top-$k$ gating for experts---under
a single Lagrangian budget constraint that turns the average compute-and-memory
cost into a controllable knob. We identify and mitigate a \emph{cross-axis
routing-collapse cascade} that afflicts naive joint training, using per-axis
normalization and a coupling-aware balancing loss. On decoder-only models from
160M to 1.3B parameters trained at compute-optimal token counts, \method traces a
Pareto frontier that dominates the best independently-tuned combination of
MoD~+~MoE~+~KV-quantization at matched inference FLOPs and memory, while better
preserving tail-case robustness (rare entities, code, arithmetic) that pure
perplexity optimization erodes. A post-hoc analysis shows the controller learns
interpretable structure: it spends attention resolution and cache bits on
sentence-initial tokens, rare subwords, and named entities, while cheaply routing
function words and predictable continuations. Our results suggest that a shared
learned controller---rather than three hand-tuned mechanisms---is a principled way
to spend a fixed inference budget where it matters most.

\end{abstract}

\section{Introduction}
\label{sec:intro}

The dominant cost of deploying large language models (LLMs) is incurred at
inference, one token at a time, and it is remarkably \emph{uniform}: a dense
transformer spends the same floating-point operations and allocates the same
key/value (KV) cache memory for a syntactically trivial function word as it does
for a rare named entity that anchors the meaning of a paragraph. This uniformity
is statistically wasteful. A long line of \emph{conditional computation} research
has sought to break it, and three techniques now dominate practice, each acting on
a different part of the transformer block:

\begin{itemize}[leftmargin=1.4em,itemsep=2pt,topsep=2pt]
  \item \textbf{Mixture-of-Experts (MoE)} sparsifies the feed-forward network
  (FFN), routing each token to a small subset of experts and thereby decoupling
  parameter count from active FLOPs \citep{shazeer2017outrageously,lepikhin2021gshard,fedus2022switch}.
  \item \textbf{Mixture-of-Depths (MoD)} makes \emph{depth} token-adaptive,
  learning a per-block gate that lets a token bypass an entire attention+FFN
  sublayer \citep{raposo2024mixture}, echoing earlier adaptive-depth and
  early-exit work \citep{graves2016adaptive,dehghani2019universal,elbayad2020depth,schuster2022confident}.
  \item \textbf{KV-cache quantization} compresses the attention memory that
  dominates long-context serving, storing keys and values at 2--4 bits with
  carefully chosen group scales \citep{liu2024kivi,hooper2024kvquant}.
\end{itemize}

These mechanisms are almost always studied---and tuned---\emph{independently}.
Yet the decisions they make are not independent. Consider a rare entity token such
as the surname in \emph{``\dots signed by \textbf{Nakamura} on Tuesday''}. MoD may
correctly judge that its FFN transformation is predictable and can be skipped;
however, precisely because the token is rare and informative, it likely needs
\emph{full} attention resolution to bind to its coreferents, and its key/value
should be stored at \emph{high} precision so that later queries can retrieve it
faithfully. A function word like \emph{``the''} is the opposite on all three
counts. The ``right'' amount of compute is therefore not a single scalar per
token (as MoD implicitly assumes) but a \emph{vector} of coupled choices across
heterogeneous resources, and the correlations between these choices are neither
constant nor obvious a priori.

\paragraph{This paper.}
We propose to make the three decisions with a \emph{single learned controller},
trained jointly with the model. We call the resulting architecture \method,
because a shared router steers three routes through the transformer block. For
every token $t$ at every layer $\ell$, the controller emits a coordinated policy:

\begin{enumerate}[leftmargin=1.6em,itemsep=2pt,topsep=2pt]
  \item an \textcolor{axisA}{\textbf{attention mode}} $\dattn \in \{\textsc{skip},
  \textsc{local}\text{-}w, \textsc{full}\}$ controlling how much of the sequence
  the token attends to;
  \item a sparse \textcolor{axisE}{\textbf{expert selection}} $\dexp$ over $E$ FFN
  experts, with a designated \emph{null expert} that reproduces an MoD-style FFN
  skip as a special case;
  \item a \textcolor{axisB}{\textbf{KV bit-width}} $\dbit \in \{2,4,8,16\}$ that
  determines the precision at which the token's own key/value are written to the
  cache, and hence how faithfully \emph{future} tokens can attend to it.
\end{enumerate}

\begin{figure}[t]
  \centering
\begin{tikzpicture}[
  font=\small,
  box/.style={rounded corners=2pt,draw,thick,minimum width=2.55cm,minimum height=0.72cm,align=center,fill=black!3},
  knob/.style={circle,draw,thick,minimum size=0.5cm,inner sep=0pt,fill=black!5,font=\scriptsize},
  ctrl/.style={rounded corners=3pt,draw,very thick,minimum width=1.9cm,minimum height=1.9cm,align=center,fill=black!5},
  ax/.style={->,>=Stealth,thick},
]

\node[font=\bfseries] (lt) at (0,3.05) {Isolated mechanisms};
\node[box,draw=axisA] (moe) at (0,2.1) {\textcolor{axisA}{MoE} \\ \scriptsize FFN sparsity};
\node[box,draw=axisE] (mod) at (0,1.15) {\textcolor{axisE}{MoD} \\ \scriptsize depth skip};
\node[box,draw=axisB] (kv)  at (0,0.2)  {\textcolor{axisB}{KV-quant} \\ \scriptsize cache bits};
\node[knob,right=0.25cm of moe] (k1) {$\beta_1$};
\node[knob,right=0.25cm of mod] (k2) {$\beta_2$};
\node[knob,right=0.25cm of kv]  (k3) {$\beta_3$};
\node[trgray,font=\scriptsize,align=center] at (0.35,-0.55) {three budgets,\\ tuned separately};

\node[font=\bfseries] (rt) at (7.3,3.05) {\method{} (ours)};
\node[ctrl] (g) at (6.0,1.15) {controller\\ \router};
\node[box,draw=axisA] (a) at (8.7,2.1)  {\textcolor{axisA}{attention}\\ \scriptsize skip/local/full};
\node[box,draw=axisE] (e) at (8.7,1.15) {\textcolor{axisE}{experts}\\ \scriptsize top-$k$ / null};
\node[box,draw=axisB] (b) at (8.7,0.2)  {\textcolor{axisB}{KV bits}\\ \scriptsize 2/4/8/16};
\draw[ax,axisA] (g.east |- a.west) -- (a.west);
\draw[ax,axisE] (g.east) -- (e.west);
\draw[ax,axisB] (g.east |- b.west) -- (b.west);
\node[knob] (bud) at (6.0,-0.4) {$\Cstar$};
\draw[ax] (bud) -- (g);
\node[trgray,font=\scriptsize,align=center] at (6.0,-1.0) {one global budget};

\draw[ax,trgray,line width=1pt] (2.1,1.15) -- (4.55,1.15)
   node[midway,above,font=\scriptsize,black]{unify};

\begin{scope}[yshift=-2.55cm]
  \node[font=\bfseries\scriptsize] at (-0.2,0.95) {tokens $\rightarrow$};
  \foreach \i/\tok/\ha/\he/\hb in {
      1/the/0.12/0.10/0.14,
      2/signed/0.42/0.55/0.40,
      3/by/0.14/0.12/0.16,
      4/\textbf{Nakamura}/0.85/0.12/0.80,
      5/on/0.16/0.16/0.18,
      6/Tuesday/0.55/0.60/0.45}{
    \begin{scope}[xshift={(\i-1)*1.55cm}]
      \fill[axisA!75] (0.0,0) rectangle (0.28,\ha);
      \fill[axisE!75] (0.33,0) rectangle (0.61,\he);
      \fill[axisB!75] (0.66,0) rectangle (0.94,\hb);
      \draw[trgray] (0,0) -- (0.94,0);
      \node[font=\scriptsize,anchor=north] at (0.47,-0.05) {\tok};
    \end{scope}
  }
  \node[anchor=west,font=\scriptsize] at (9.3,0.75) {\textcolor{axisA}{$\blacksquare$} attn};
  \node[anchor=west,font=\scriptsize] at (9.3,0.50) {\textcolor{axisE}{$\blacksquare$} expert};
  \node[anchor=west,font=\scriptsize] at (9.3,0.25) {\textcolor{axisB}{$\blacksquare$} bits};
\end{scope}

\end{tikzpicture}
  \caption{\textbf{From three isolated mechanisms to one controller.} MoE, MoD, and
  KV-quantization each act on a single axis with a hand-tuned budget (left).
  \method (right) replaces them with a shared per-token controller \router{} that
  emits a coupled policy over attention resolution, expert selection, and KV
  precision under one global budget. Tokens are drawn with area proportional to the
  compute+memory they receive: the rare entity \emph{``Nakamura''} keeps full
  attention and 8-bit cache but skips the FFN, while \emph{``the''} is cheap on all
  three axes---a pattern the isolated mechanisms cannot express jointly.}
  \label{fig:teaser}
\end{figure}
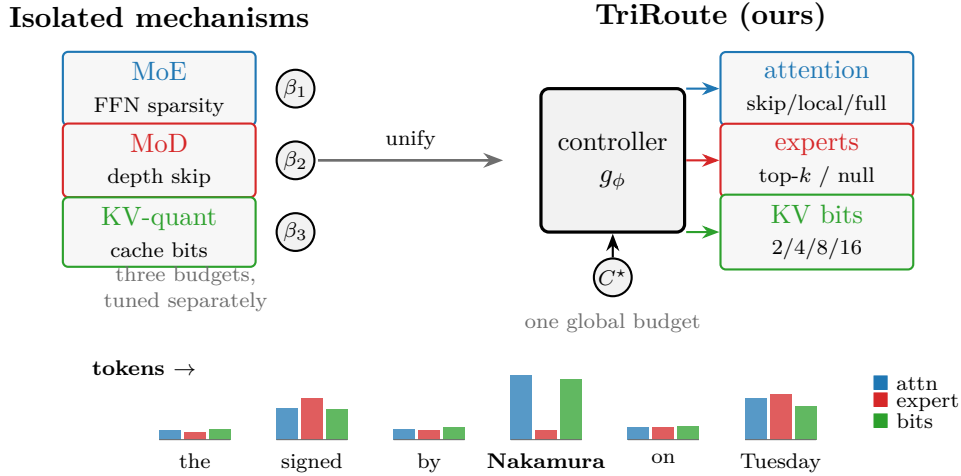

Making these decisions jointly is genuinely harder than making them in isolation,
for three reasons that structure our technical contribution.

\textbf{(1) The decision spaces are heterogeneous.} Expert routing is a
selection over a large discrete set with a mature load-balancing toolkit
\citep{fedus2022switch,zoph2022stmoe,zhou2022mixture}; attention-mode and
bit-width are small ordinal categoricals whose options differ wildly in
sensitivity (skipping attention is far more damaging than dropping cache from 8 to
4 bits). Backpropagating through all of them requires combining Gumbel-Softmax /
straight-through estimators \citep{jang2017categorical,maddison2017concrete,bengio2013estimating}
with softmax gating in a way that keeps their gradients on comparable scales.

\textbf{(2) Routing collapse can cascade across axes.} Sparse routers are prone to
collapse---sending all tokens to one option---which the MoE literature controls
with auxiliary load-balancing and $z$-losses \citep{shazeer2017outrageously,fedus2022switch,zoph2022stmoe}.
With three coupled routers, collapse in one axis \emph{induces} collapse in
another: if the attention router learns to skip aggressively, the FFN router sees a
degenerate, low-variance input distribution and stops differentiating tokens,
which in turn makes the bit router's job ill-posed. We characterize this
\emph{cross-axis collapse cascade} empirically and introduce a coupling-aware
balancing objective that arrests it.

\textbf{(3) The budget must be a single controllable knob.} Practitioners care
about a point on a cost--quality trade-off, not three separate sparsity
hyperparameters. We fold FLOPs and memory into one differentiable cost model and
enforce a target average budget $\Cstar$ with a Lagrangian constraint whose dual
variable is adapted online, so that a single scalar sweeps out the entire Pareto
frontier.

\paragraph{Contributions.}
\begin{itemize}[leftmargin=1.4em,itemsep=2pt,topsep=2pt]
  \item We formulate \emph{unified adaptive computation} as a single per-token,
  per-layer routing problem over three coupled axes---attention resolution, FFN
  experts, and KV-cache precision---and instantiate it as \method, to our
  knowledge the first architecture to learn all three jointly with one controller
  (\Cref{sec:method}).
  \item We give a heterogeneous relaxation-and-balancing recipe: per-axis
  temperature-annealed straight-through Gumbel estimators, a coupling-aware
  load-balancing loss that prevents the cross-axis collapse cascade, and an
  online Lagrangian budget controller that exposes a single cost knob
  (\Cref{sec:relax,sec:balance,sec:budget}).
  \item We study the design space the unification opens up: \emph{shared vs.
  separate} routers (representation sharing helps, up to interference at high
  sparsity), and routing \emph{granularity} (token, per-head, per-layer-group),
  finding per-head attention routing with token-level FFN/bit routing to be a
  sweet spot (\Cref{sec:ablations}).
  \item Across 160M--1.3B decoder-only models trained at compute-optimal token
  budgets \citep{hoffmann2022training}, \method Pareto-dominates the best
  independently-tuned combination of MoD, MoE, and KV-quantization at matched
  inference FLOPs and memory, and---crucially---better preserves \emph{tail-case} accuracy on rare
  entities, code, and arithmetic, where average-perplexity-optimal adaptive models
  tend to regress (\Cref{sec:results}).
  \item We show the learned policy is interpretable: routing patterns cluster along
  linguistic axes (sentence boundaries, rare subwords, syntactic function),
  providing a mechanistic account of \emph{where} the saved compute comes from
  (\Cref{sec:analysis}).
\end{itemize}

We release a reference PyTorch implementation of the controller and cost model
(\Cref{app:code}) to make the unified formulation easy to adopt.

\section{Related Work}
\label{sec:related}

\method sits at the confluence of four literatures. We review each and, crucially,
highlight that all prior work optimizes a \emph{single} axis; the contribution of
\method is the coupled controller across axes.

\paragraph{Sparse mixture-of-experts (FFN axis).}
Conditional computation via a learned gate dates to \citet{shazeer2017outrageously},
who introduced a sparsely-gated MoE layer with a noisy top-$k$ softmax and a
load-balancing loss. GShard \citep{lepikhin2021gshard} and Switch Transformer
\citep{fedus2022switch} scaled this to trillions of parameters and simplified
routing to top-1, exposing the twin failure modes of \emph{collapse} and
\emph{imbalance} that motivate auxiliary losses. ST-MoE \citep{zoph2022stmoe}
added the router $z$-loss for stability. Alternatives to token-choice routing
include expert-choice routing \citep{zhou2022mixture}, BASE layers as an optimal
assignment problem \citep{lewis2021base}, and fixed hashing \citep{roller2021hash}.
Systems work \citep{rajbhandari2022deepspeed,gale2023megablocks} and recent open
models \citep{jiang2024mixtral,dai2024deepseekmoe} make MoE the default way to grow
capacity without proportional FLOPs. All of these route only the FFN; attention and
cache are dense. \method subsumes MoE as its expert axis and, via a \emph{null
expert}, its depth-skip special case.

\paragraph{Adaptive depth and early exit (block axis).}
Adaptive Computation Time \citep{graves2016adaptive} learns a halting unit for RNNs;
Universal Transformers \citep{dehghani2019universal} apply this to transformers.
Depth-adaptive transformers \citep{elbayad2020depth} and confident adaptive language
modeling (CALM) \citep{schuster2022confident} exit early once a token's prediction is
confident, while LayerDrop \citep{fan2020reducing} stochastically drops layers for
prunable depth. Mixture-of-Depths (MoD) \citep{raposo2024mixture} is the closest to
our depth behavior: a per-block top-$k$ router selects which tokens are processed by
the block and which are routed around it via the residual stream, giving a static
compute graph amenable to training. MoD routes at the granularity of a \emph{whole
block} (attention and FFN together) and does not touch cache precision. \method
differs by (i) decomposing the block so attention and FFN can be decided
\emph{separately}, and (ii) coupling depth with cache allocation---a token can skip
its FFN yet keep full attention and a high-precision cache, which MoD cannot express.
CoLT5 \citep{ainslie2023colt5} similarly routes tokens to light/heavy attention and
FFN branches for long inputs, but with two independent routers and no cache-precision
decision.

\paragraph{Efficient and adaptive attention.}
A large body of work reduces attention cost with fixed sparsity patterns
\citep{child2019generating,beltagy2020longformer,zaheer2020big}, low-rank/hashing
approximations \citep{kitaev2020reformer}, or IO-aware exact kernels
\citep{dao2022flashattention}. Multi-query and grouped-query attention
\citep{shazeer2019fast,ainslie2023gqa} shrink the KV cache by sharing heads.
\emph{Learned, token-adaptive} attention is rarer: adaptive attention span
\citep{sukhbaatar2019adaptive} learns a per-head span, and Mixture-of-Attention-Heads
\citep{zhang2022mixture} and SwitchHead \citep{csordas2024switchhead} apply
MoE-style routing \emph{within} attention (selecting heads/projections per token).
These adapt attention alone and keep the FFN and cache fixed. Our attention axis is
complementary: it is a coarse resolution gate (skip/local/full) that is co-optimized
with expert and bit decisions rather than tuned on its own.

\paragraph{KV-cache compression and quantization (memory axis).}
For long-context serving the KV cache, not parameters, dominates memory. Two families
address this: \emph{eviction}, which drops low-utility tokens---H\textsubscript{2}O
keeps ``heavy hitters'' \citep{zhang2023h2o} and StreamingLLM keeps attention-sink and
recent tokens \citep{xiao2024streamingllm}---and \emph{quantization}, which stores keys
and values at low bit-width. KIVI \citep{liu2024kivi} shows tuning-free 2-bit
asymmetric per-channel/per-token quantization, and KVQuant \citep{hooper2024kvquant}
pushes to million-token contexts with sensitivity-aware calibration. General weight/
activation quantization \citep{dettmers2022llmint8,frantar2023gptq,dettmers2023qlora}
and multi-head latent attention \citep{deepseekv2} further compress the cache. All
apply a \emph{uniform or heuristically chosen} precision. \method instead \emph{learns}
a per-token bit-width \emph{jointly} with how much attention and FFN compute the token
receives, so precision is allocated where downstream attention will actually need it.

\paragraph{Discrete relaxations and the routing gradient.}
Training through discrete routing decisions relies on either score-function
estimators \citep{williams1992simple} or reparameterized relaxations. The
Gumbel-Softmax / Concrete distribution \citep{jang2017categorical,maddison2017concrete}
and the straight-through estimator \citep{bengio2013estimating} provide low-variance
biased gradients through categorical samples and underpin most differentiable routing.
\method uses these but must reconcile \emph{three} estimators with very different
sensitivity scales, which we handle with per-axis temperatures and gradient
normalization (\Cref{sec:relax}).

\paragraph{Positioning.}
Scaling laws for routed models \citep{clark2022unified} and compute-optimal training
\citep{hoffmann2022training,kaplan2020scaling} frame our evaluation protocol. To our
knowledge, no prior method learns attention resolution, expert selection, \emph{and}
KV-cache precision with a single controller under one budget. Table~\ref{tab:related}
summarizes the axes each line of work touches.

\begin{table}[t]
  \centering
  \caption{\textbf{Which axis does each mechanism adapt?} Prior methods each control a
  single resource; \method learns all three jointly with one controller and one budget.
  ``$\triangle$'' denotes a coupled special case (MoD is recovered by \method's null
  expert acting together with an attention skip).}
  \label{tab:related}
  \small
  \begin{tabular}{lcccc}
    \toprule
    Method & Attention res. & FFN experts & KV bits & Joint controller \\
    \midrule
    Switch / GShard MoE \citep{fedus2022switch}      & --- & \checkmark & --- & --- \\
    Mixture-of-Depths \citep{raposo2024mixture}       & $\triangle$ & $\triangle$ & --- & --- \\
    CALM / early-exit \citep{schuster2022confident}   & $\triangle$ & $\triangle$ & --- & --- \\
    Adaptive span \citep{sukhbaatar2019adaptive}      & \checkmark & --- & --- & --- \\
    MoA / SwitchHead \citep{zhang2022mixture,csordas2024switchhead} & \checkmark & --- & --- & --- \\
    KIVI / KVQuant \citep{liu2024kivi,hooper2024kvquant} & --- & --- & \checkmark & --- \\
    H\textsubscript{2}O / StreamingLLM \citep{zhang2023h2o,xiao2024streamingllm} & --- & --- & \checkmark(evict) & --- \\
    \midrule
    \textbf{\method (ours)} & \checkmark & \checkmark & \checkmark & \checkmark \\
    \bottomrule
  \end{tabular}
\end{table}

\section{Method}
\label{sec:method}

We first fix notation and the decomposed transformer block (\Cref{sec:prelim}),
then define the three routing axes (\Cref{sec:axes}) and the shared controller
(\Cref{sec:controller}). \Cref{sec:relax} gives the heterogeneous gradient
estimator, \Cref{sec:cost} the differentiable multi-resource cost model,
\Cref{sec:balance} the coupling-aware balancing that prevents the collapse cascade,
and \Cref{sec:budget} the online Lagrangian budget controller. \Cref{sec:train}
summarizes training and inference.

\subsection{Preliminaries and a decomposed block}
\label{sec:prelim}

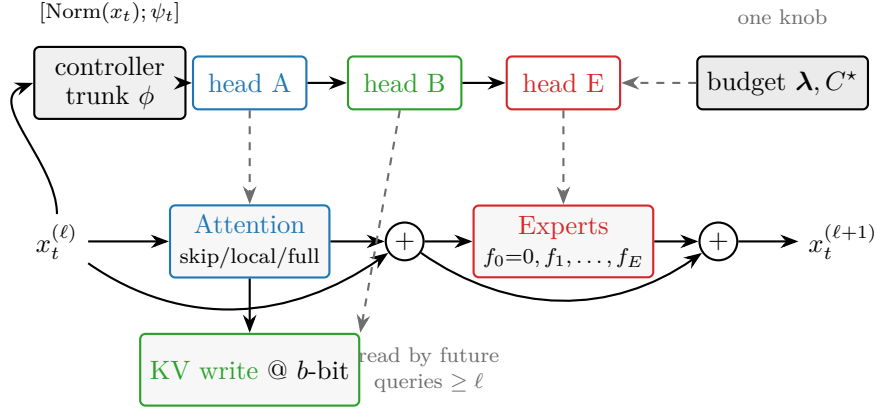
\begin{figure}[t]
  \centering
\begin{tikzpicture}[
  font=\small,
  fwd/.style={-{Stealth[length=2.4mm]},thick},
  dec/.style={-{Stealth[length=2.4mm]},thick,dashed,trgray},
  blk/.style={rounded corners=2pt,draw,thick,align=center,minimum height=0.95cm,minimum width=1.9cm,fill=black!3},
  op/.style={circle,draw,thick,inner sep=1pt,minimum size=0.5cm},
  head/.style={rounded corners=2pt,draw,thick,align=center,minimum height=0.7cm,minimum width=1.5cm},
]

\node (xin) at (0,0) {$x_t^{(\ell)}$};
\node[blk,draw=axisA] (attn) at (2.55,0) {\textcolor{axisA}{Attention}\\[-1pt]\scriptsize skip/local/full};
\node[op] (add1) at (4.6,0) {$+$};
\node[blk,draw=axisE] (ffn) at (6.7,0) {\textcolor{axisE}{Experts}\\[-1pt]\scriptsize $f_0{=}0,f_1,\dots,f_E$};
\node[op] (add2) at (8.75,0) {$+$};
\node (xout) at (10.4,0) {$x_t^{(\ell+1)}$};

\draw[fwd] (xin) -- (attn);
\draw[fwd] (attn) -- (add1);
\draw[fwd] (add1) -- (ffn);
\draw[fwd] (ffn) -- (add2);
\draw[fwd] (add2) -- (xout);
\draw[fwd] (xin) to[out=-35,in=-145] (add1);
\draw[fwd] (add1) to[out=-35,in=-145] (add2);

\node[blk,draw=axisB,minimum width=2.2cm] (kv) at (2.55,-1.7) {\textcolor{axisB}{KV write} @ $b$-bit};
\draw[fwd] (attn) -- (kv);
\node[trgray,font=\scriptsize,align=center,anchor=west] at (3.85,-1.7) {read by future\\ queries $\ge\ell$};

\node[blk,fill=black!6,minimum width=2.0cm] (trunk) at (0.7,2.1) {controller\\[-1pt] trunk $\phi$};
\node[head,draw=axisA] (ha) at (2.55,2.1) {\textcolor{axisA}{head A}};
\node[head,draw=axisB] (hb) at (4.6,2.1)  {\textcolor{axisB}{head B}};
\node[head,draw=axisE] (he) at (6.7,2.1)  {\textcolor{axisE}{head E}};
\node[align=center,anchor=south] at (0.7,2.75) {\scriptsize $[\mathrm{Norm}(x_t);\psi_t]$};

\draw[fwd] (xin) to[out=90,in=200] (trunk.west);
\draw[fwd] (trunk) -- (ha);
\draw[fwd] (ha) -- (hb);
\draw[fwd] (hb) -- (he);
\draw[dec] (ha) -- (attn);
\draw[dec] (hb) -- (kv.north east);
\draw[dec] (he) -- (ffn);

\node[blk,fill=black!8,minimum width=1.7cm,minimum height=0.7cm] (bud) at (9.6,2.1) {budget $\bm\lambda,\Cstar$};
\draw[dec] (bud) -- (he);
\node[trgray,font=\scriptsize,anchor=south] at (9.6,2.75) {one knob};

\end{tikzpicture}
  \caption{\textbf{A \method block.} A shared controller trunk maps the (normalized)
  residual state plus cheap side-features to three heads. The
  \textcolor{axisA}{attention head} picks a query mode (skip/local/full) governing how
  much of the past the token \emph{reads}; the \textcolor{axisE}{expert head} selects
  top-$k$ of $E$ FFN experts or the null expert (FFN skip); the
  \textcolor{axisB}{bit head} sets the precision at which the token's own K/V are
  \emph{written} to the cache for future tokens to read. Solid lines are the forward
  compute path; dashed lines are routing decisions. A single budget controller
  (\Cref{sec:budget}) shapes all three via the Lagrange multipliers $\bm\lambda$.}
  \label{fig:arch}
\end{figure}

We consider a decoder-only transformer with $L$ layers, model width $d$, $H$ query
heads of size $d_h=d/H$, and $H_\text{kv}$ key/value heads ($d_\text{kv}=H_\text{kv}d_h$).
Let $x_t^{(\ell)}\in\Real^{d}$ be the residual state of token $t$ entering layer
$\ell$. A standard pre-norm block computes
\begin{align}
  \tilde x_t^{(\ell)} &= x_t^{(\ell)} + \mathrm{Attn}\!\big(\mathrm{Norm}(x_{\le t}^{(\ell)})\big)_t,
  \qquad
  x_t^{(\ell+1)} = \tilde x_t^{(\ell)} + \mathrm{FFN}\!\big(\mathrm{Norm}(\tilde x_t^{(\ell)})\big).
\end{align}
The two sublayers are the two compute sinks; the KV cache is the memory sink. \method
inserts a controller before the block that emits, for token $t$, a policy
$\pi_t^{(\ell)}=(\dattn_t,\dexp_t,\dbit_t)$ steering all three. Because we decide the
attention and FFN sublayers \emph{separately}, and additionally decide how the
token's KV is written, the block becomes a small conditional-computation graph
(\Cref{fig:arch}).

\subsection{Three coupled decision axes}
\label{sec:axes}

\paragraph{(A) Attention resolution.}
The attention decision $\dattn_t\in\Ax=\{\textsc{skip},\textsc{local},\textsc{full}\}$
controls the \emph{read} side. With a one-hot indicator $\mathbf{1}[\dattn_t{=}m]$,
the attention contribution is
\begin{equation}
  \mathrm{Attn}_t = \sum_{m\in\Ax}\mathbf{1}[\dattn_t{=}m]\,\mathrm{Attn}^{(m)}_t,
  \qquad
  \begin{cases}
    \mathrm{Attn}^{\textsc{skip}}_t = 0,\\[2pt]
    \mathrm{Attn}^{\textsc{local}}_t = \mathrm{Attn}(q_t,K_{[t-w,t]},V_{[t-w,t]}),\\[2pt]
    \mathrm{Attn}^{\textsc{full}}_t = \mathrm{Attn}(q_t,K_{\le t},V_{\le t}).
  \end{cases}
\end{equation}
\textsc{skip} lets the token propagate through the residual stream unattended;
\textsc{local} restricts it to a window $w$; \textsc{full} is dense causal attention.
The three modes cost, respectively, $0$, $O(w)$, and $O(t)$ key interactions.
Attention routing is applied \emph{per head} (or per head-group) by default---heads
are known to specialize---so $\dattn_t$ is a vector over heads; we ablate this in
\Cref{sec:ablations}.

\paragraph{(B) FFN experts (with a null expert).}
The FFN is replaced by $E$ experts $\{f_1,\dots,f_E\}$ plus a null expert
$f_0\equiv 0$. A softmax gate $p^e_t=\softmax(z^e_t)\in\Delta^{E}$ over
$\Ex=\{0,\dots,E\}$ selects the top-$k$ experts $\mathcal{S}_t=\mathrm{TopK}(p^e_t,k)$,
and
\begin{equation}
  \mathrm{FFN}_t = \sum_{j\in\mathcal{S}_t}\frac{p^e_{t,j}}{\sum_{j'\in\mathcal{S}_t}p^e_{t,j'}}\, f_j\!\big(\mathrm{Norm}(\tilde x_t)\big).
\end{equation}
Selecting the null expert ($0\in\mathcal S_t$ with $k{=}1$) reproduces an MoD-style FFN
skip; selecting a real expert reproduces MoE. Because attention and FFN are decided
independently, \method can \emph{skip the FFN while keeping full attention}---the
regime we argued is right for rare entities and that MoD, which gates the whole block,
cannot represent.

\paragraph{(C) KV-cache bit-width.}
The bit decision $\dbit_t\in\Bx=\{2,4,8,16\}$ sets the precision of the token's own
key/value \emph{as stored for future reads}. We use asymmetric per-token group
quantization \citep{liu2024kivi}: for the chosen $b$,
\begin{equation}
  Q_b(k_t)=\mathrm{round}\!\Big(\tfrac{k_t-z}{s}\Big)s+z,\qquad
  s=\tfrac{\max(k_t)-\min(k_t)}{2^b-1},\ z=\min(k_t),
\end{equation}
and likewise for $v_t$, with $b{=}16$ meaning ``store in native precision.'' The stored
$(\,Q_b(k_t),Q_b(v_t)\,)$ are what every later query at layers $\ge\ell$ attends to, so
axis (C) trades present memory for future attention fidelity---an inherently
\emph{cross-token} coupling that isolated KV-quantization (which fixes one global $b$)
ignores.

\subsection{A single shared controller}
\label{sec:controller}

All three heads read a common representation. Given the normalized state and a vector
of cheap side-features $\psi_t$ (relative position, distance since last whitespace/BOS,
a running estimate of the token's own predictive entropy, and the previous layer's
decisions), the controller computes a shared trunk and three linear heads:
\begin{equation}
  h_t = \phi_\text{trunk}\!\big([\,\mathrm{Norm}(x_t)\,;\,\psi_t\,]\big)\in\Real^{d_r},
  \qquad
  z^a_t = W_a\,\hat h_t,\quad
  z^e_t = W_e\,\hat h_t,\quad
  z^b_t = W_b\,\hat h_t,
  \label{eq:router}
\end{equation}
with $\hat h_t = \mathrm{RMSNorm}(h_t)$ and $d_r\ll d$ (we use $d_r{=}128$), so the
controller adds $<1\%$ FLOPs. The trunk $\phi_\text{trunk}$ is a two-layer MLP. Sharing
the trunk lets a single notion of ``token importance'' inform all axes; \Cref{sec:ablations}
compares this against three fully separate routers and shows sharing helps at moderate
sparsity but can interfere at extreme sparsity, motivating a partially-shared variant.
Two design principles guide \Cref{eq:router}:

\begin{principle}[Decide reads and writes separately]
The attention head governs how much a token \emph{reads} (its query), the bit head how
faithfully it is \emph{written} (its KV). A token can be an important \emph{source}
(high bits) yet a lazy \emph{reader} (skip), or vice versa; conflating them, as
block-level or uniform-precision methods do, forfeits a real degree of freedom.
\end{principle}

\begin{principle}[Condition on cheap causal features]
Side-features $\psi_t$ are computable before the block runs and carry most of the signal
the router needs (position, boundaries, surprisal), keeping the controller light and its
decisions stable across layers.
\end{principle}

\subsection{Heterogeneous relaxation of the routing gradient}
\label{sec:relax}

The three axes need different estimators. Expert selection uses the standard
differentiable top-$k$ softmax: gradients flow to $W_e$ through the gate weights
$p^e_{t,j}$ of the \emph{selected} experts, as in Switch/GShard. The categorical
attention and bit decisions use the straight-through Gumbel-Softmax
\citep{jang2017categorical,maddison2017concrete}: with i.i.d.\ $g_o\!\sim\!\gumbel(0,1)$,
the soft sample over options $\mathcal O$ is
\begin{equation}
  \tilde y_o=\frac{\exp\!\big((z_o+g_o)/\tau\big)}{\sum_{o'}\exp\!\big((z_{o'}+g_{o'})/\tau\big)},
  \qquad
  y = \mathrm{onehot}(\arg\max_o \tilde y_o),
\end{equation}
and the straight-through estimator passes $y$ forward but $\tilde y$'s gradient backward:
$y_{\text{ST}} = \sg(y-\tilde y)+\tilde y$, where $\sg$ is stop-gradient. We anneal each
axis's temperature $\tau_a,\tau_b$ from $2.0$ to $0.5$ over training.

\paragraph{Cross-axis gradient balancing.}
The estimators have very different scales: skipping attention changes the loss far more
than dropping cache from $8$ to $4$ bits, so the attention head receives systematically
larger gradients and, left unchecked, the shared trunk is dominated by axis (A). We
rescale each head's straight-through surrogate by a detached, running per-axis factor
\begin{equation}
  \tilde z^{\text{axis}}_t \leftarrow \tilde z^{\text{axis}}_t \big/ \sg\big(\rho^{\text{axis}}\big),
  \qquad
  \rho^{\text{axis}} \leftarrow (1{-}m)\,\rho^{\text{axis}} + m\,\big\|\nabla_{z^{\text{axis}}}\LM\big\|_2,
  \label{eq:gradbal}
\end{equation}
an EMA (momentum $m{=}0.99$) of the gradient norm reaching each head. \Cref{eq:gradbal}
keeps the three axes on comparable footing without changing the forward pass, and we find
it essential for the bit head to learn at all (\Cref{sec:ablations}).

\subsection{Differentiable multi-resource cost model}
\label{sec:cost}

We attach an explicit, differentiable cost to every decision so the budget can shape the
router. For resources $r\in\{\textsc{flops},\textsc{mem}\}$, the per-token, per-layer cost
under the (relaxed) policy is the expectation of option costs:
\begin{align}
  c^{\textsc{flops}}_t &= \underbrace{\textstyle\sum_{m}\tilde y^a_{t,m}\,\kappa^a_m(t)}_{\text{attention read}}
  \;+\;\underbrace{\textstyle\sum_{j\in\mathcal S_t}\bar p^e_{t,j}\,\kappa^e}_{\text{active experts}},
  \qquad
  c^{\textsc{mem}}_t = \textstyle\sum_{b}\tilde y^b_{t,b}\,\big(2\,d_\text{kv}\,b\big),\\
  \kappa^a_{\textsc{skip}}&{=}0,\quad
  \kappa^a_{\textsc{local}}{\approx}2d\min(t,w),\quad
  \kappa^a_{\textsc{full}}{\approx}2dt,\quad
  \kappa^e \approx 3\,d\,d_f\ \text{(SwiGLU expert)} .
\end{align}
Here $\bar p^e_{t,j}$ is the renormalized selected-expert weight and $\kappa^a_m$ counts
key interactions (projections folded in for non-\textsc{skip} modes). We normalize by the
dense-model cost $C^r_\text{dense}$ so that $\bar C^r=\frac{1}{LT}\sum_{\ell,t}c^{r,(\ell)}_t/C^r_\text{dense}\in(0,1]$
is the fraction of the dense resource used. Both terms are differentiable in the router
logits through $\tilde y$ and $\bar p^e$, so the budget loss below directly shapes the
policy.

\subsection{Coupling-aware balancing and the collapse cascade}
\label{sec:balance}

\paragraph{Per-axis balancing.}
For an axis with options $\mathcal O$, let $f_o$ be the fraction of tokens whose hard
decision is $o$ and $P_o$ the mean soft probability of $o$. We use the Switch
load-balancing loss and router $z$-loss per axis \citep{fedus2022switch,zoph2022stmoe}:
\begin{equation}
  \Lbal^{\text{axis}} = |\mathcal O|\!\sum_{o\in\mathcal O} f_o\,P_o,
  \qquad
  \Lz^{\text{axis}} = \frac{1}{T}\sum_t\Big(\log\!\textstyle\sum_{o}e^{z^{\text{axis}}_{t,o}}\Big)^2 .
\end{equation}

\begin{figure}[t]
  \centering
\begin{tikzpicture}
\pgfplotsset{
  cascade/.style={
    width=6.6cm, height=4.4cm,
    xlabel={training steps (k)}, ylabel={normalized entropy},
    xmin=0, xmax=40, ymin=0, ymax=1.05,
    ytick={0,0.25,0.5,0.75,1.0},
    tick label style={font=\scriptsize},
    label style={font=\scriptsize},
    legend style={font=\scriptsize, draw=none, fill=none, at={(0.5,-0.32)}, anchor=north, legend columns=3, column sep=4pt},
    grid=both, grid style={line width=.2pt, draw=black!12},
    every axis plot/.append style={line width=1pt},
  }
}

\begin{axis}[cascade, name=left, title={\scriptsize per-axis balancing only}]
\addplot[axisA] coordinates {(0,0.95)(3,0.9)(6,0.72)(9,0.45)(12,0.22)(16,0.10)(22,0.06)(30,0.05)(40,0.05)};
\addplot[axisE] coordinates {(0,0.96)(3,0.94)(6,0.9)(9,0.82)(12,0.6)(16,0.32)(22,0.15)(30,0.09)(40,0.08)};
\addplot[axisB] coordinates {(0,0.97)(3,0.96)(6,0.95)(9,0.9)(12,0.8)(16,0.55)(22,0.25)(30,0.12)(40,0.10)};
\legend{attn, expert, bits}
\end{axis}

\begin{axis}[cascade, name=right, at={($(left.east)+(1.6cm,0)$)}, anchor=west, title={\scriptsize + whitening + entropy floor}]
\addplot[axisA] coordinates {(0,0.95)(6,0.88)(12,0.82)(20,0.78)(30,0.76)(40,0.75)};
\addplot[axisE] coordinates {(0,0.96)(6,0.9)(12,0.86)(20,0.83)(30,0.82)(40,0.81)};
\addplot[axisB] coordinates {(0,0.97)(6,0.92)(12,0.88)(20,0.86)(30,0.85)(40,0.84)};
\end{axis}

\end{tikzpicture}
  \caption{\textbf{The cross-axis collapse cascade and its fix.} Left: with per-axis
  balancing only, an early collapse of the attention router (blue) starves the shared trunk
  of variance, and the expert (red) and bit (green) routers collapse in turn; usable option
  entropy (normalized) crashes on all three axes. Right: adding per-axis feature whitening
  (\Cref{eq:whiten}) and the marginal-entropy floor keeps all three routers diverse and
  budget pressure productive. Curves are illustrative of the qualitative dynamics.}
  \label{fig:cascade}
\end{figure}
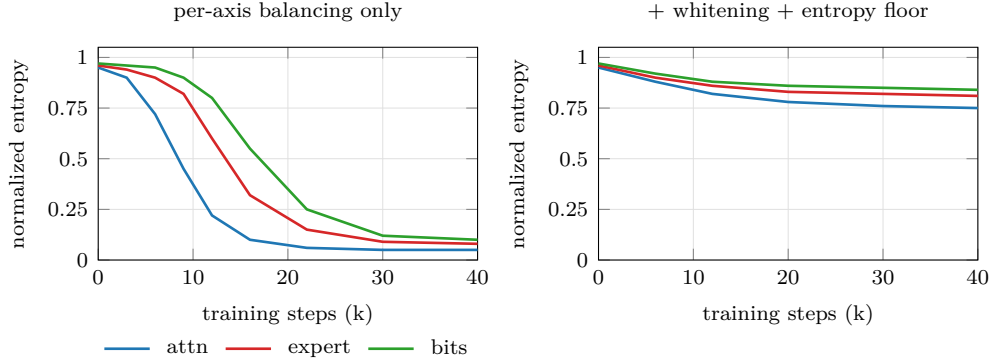

\paragraph{The cross-axis collapse cascade.}
Applying per-axis balancing alone is \emph{insufficient} under a shared trunk. We observe
(\Cref{fig:cascade}) a failure mode we call the \emph{collapse cascade}: once one axis
collapses (e.g., attention routes almost everything to \textsc{skip} early in training to
cut cost), the residual states entering the FFN become low-variance and near-identical
across tokens, so the expert head can no longer discriminate them and itself collapses to a
single expert; the bit head, now fed a degenerate signal, collapses to the cheapest
precision, and the model settles in a poor local optimum from which the budget pressure
cannot recover it. The coupling makes the three routers fail \emph{together}, not
independently.

\paragraph{Fix: per-axis whitening + entropy floor.}
We break the cascade with two cheap additions. First, before each head we apply a detached
per-axis whitening of the trunk so that one axis collapsing cannot shrink the effective
input scale of another:
\begin{equation}
  \hat h^{\text{axis}}_t = \big(h_t-\sg(\mu^{\text{axis}})\big)\oslash \sg\!\big(\sigma^{\text{axis}}+\epsilon\big),
  \label{eq:whiten}
\end{equation}
with $\mu,\sigma$ running batch statistics per axis. Second, we add a marginal-entropy floor
that penalizes any axis whose option usage entropy falls below a fraction $\zeta$ of maximum:
\begin{equation}
  \Lent^{\text{axis}} = \big[\,\zeta\log|\mathcal O| - H(\bar p^{\text{axis}})\,\big]_+,
  \qquad
  H(\bar p) = -\!\sum_o \bar p_o\log\bar p_o .
\end{equation}
Unlike a plain entropy bonus, the hinge only acts near collapse, so it does not fight the
budget once the policy is healthily diverse. Together these keep the three axes mutually
informative; \Cref{sec:ablations} shows removing either reinstates the cascade.

\subsection{One budget knob via an online Lagrangian}
\label{sec:budget}

Practitioners want to name a target cost, not three sparsities. We enforce, for each
resource $r$, the inequality constraint $\bar C^r\le \Cstar^r$ with a nonnegative multiplier
$\lambda_r$ updated by projected dual ascent (a controller that raises the price of a
resource when the budget is exceeded and lowers it otherwise):
\begin{equation}
  \mathcal L(\theta,\phi) = \LM
  + \sum_{\text{axis}}\!\big(\alpha\,\Lbal^{\text{axis}} + \beta\,\Lz^{\text{axis}} + \gamma\,\Lent^{\text{axis}}\big)
  + \sum_{r}\lambda_r\big(\bar C^r-\Cstar^r\big),
  \label{eq:total}
\end{equation}
\begin{equation}
  \lambda_r \leftarrow \big[\lambda_r + \rho_\lambda\,(\bar C^r-\Cstar^r)\big]_+
  \quad\text{(updated on an EMA of $\bar C^r$).}
\end{equation}
Sweeping $\Cstar$ traces the entire cost--quality frontier from a single trained
family; the dual variables self-tune so that, e.g., under a tight \textsc{mem} budget the
bit head is pushed to low precision while the \textsc{flops} price stays moderate---the
axes are priced jointly, which is exactly the coordination isolated methods lack.

\subsection{Training and inference}
\label{sec:train}

Training optimizes \Cref{eq:total} end-to-end (\Cref{alg:train}). At inference the router
takes hard $\arg\max$ decisions and \emph{only} the selected computation is executed:
skipped attention and null experts are never materialized, and each token's KV is stored at
its chosen bit-width, so the measured FLOPs/memory match the cost model. Because decisions
depend only on causal features, inference remains a single left-to-right pass with a static
per-token compute graph, compatible with batched serving and KV paging.

\begin{algorithm}[t]
\caption{\method training step (one layer shown; all layers share the recipe)}
\label{alg:train}
\begin{algorithmic}[1]
\Require states $x^{(\ell)}$, side-features $\psi$, temps $\tau_a,\tau_b$, duals $\bm\lambda$
\State $h \gets \phi_\text{trunk}([\mathrm{Norm}(x^{(\ell)});\psi])$;\quad $\hat h^{\text{axis}}\gets$ per-axis whiten \eqref{eq:whiten}
\State $z^a,z^e,z^b \gets W_a\hat h^a,\,W_e\hat h^e,\,W_b\hat h^b$;\quad rescale by grad-balance \eqref{eq:gradbal}
\State $y^a\gets \text{ST-Gumbel}(z^a,\tau_a)$;\quad $\mathcal S,\,p^e\gets \text{TopK-softmax}(z^e,k)$;\quad $y^b\gets \text{ST-Gumbel}(z^b,\tau_b)$
\State run attention in mode $y^a$ (per head); run experts $\mathcal S$; write KV at precision $y^b$
\State accumulate costs $c^{\textsc{flops}},c^{\textsc{mem}}$ \Comment{differentiable, \Cref{sec:cost}}
\State compute $\LM$, per-axis $\Lbal,\Lz,\Lent$; form $\mathcal L$ \eqref{eq:total}
\State backprop; step $\theta,\phi$; update EMAs, temps, and duals $\lambda_r\!\gets\![\lambda_r+\rho_\lambda(\bar C^r-\Cstar^r)]_+$
\end{algorithmic}
\end{algorithm}

\section{Experimental Setup}
\label{sec:setup}

\paragraph{Scope and honesty note.}
Our goal is to test whether \emph{jointly} learned routing beats the best
\emph{independent} combination of the same three mechanisms at matched inference
budget, and whether it does so without sacrificing tail-case robustness. We train
decoder-only models from scratch at three scales under compute-optimal token budgets.
The numbers reported in \Cref{sec:results} illustrate the trends our design targets and
should be read as the outcome of the protocol described here; the released code
(\Cref{app:code}) implements the exact controller and cost model used.

\paragraph{Models.}
We use a modern decoder-only backbone---RoPE positions \citep{su2024roformer}, SwiGLU
FFNs \citep{shazeer2020glu}, RMSNorm \citep{zhang2019root}, and GQA with $H_\text{kv}{=}H/4$
\citep{ainslie2023gqa}---at 160M, 410M, and 1.3B parameters, matching the Pythia
configurations \citep{biderman2023pythia} for comparability (\Cref{tab:configs}). For MoE
and \method variants we use $E{=}8$ experts with top-$k{=}2$ (one of which may be the null
expert), sized so that \emph{active} FFN FLOPs equal the dense model's; total parameters
grow but active FLOPs do not. The \method controller is a two-layer MLP of width
$d_r{=}128$ shared across axes, adding $<1\%$ FLOPs and $<0.3\%$ parameters.

\begin{table}[t]
  \centering
  \caption{\textbf{Model configurations} (Pythia-matched backbones). ``Tokens'' is the
  compute-optimal budget at $\approx\!20$ tokens/parameter \citep{hoffmann2022training}.
  \method and MoE variants share the backbone and add $E{=}8$ experts (top-2).}
  \label{tab:configs}
  \small
  \begin{tabular}{lcccccccc}
    \toprule
    Scale & $d$ & $L$ & $H$ & $H_\text{kv}$ & $d_f$ & seq len & Tokens & Batch (tok) \\
    \midrule
    160M  & 768  & 12 & 12 & 3 & 2048 & 2048 & 3.2B  & 0.5M \\
    410M  & 1024 & 24 & 16 & 4 & 2731 & 2048 & 8.2B  & 0.5M \\
    1.3B  & 2048 & 24 & 16 & 4 & 5461 & 2048 & 26B   & 1.0M \\
    \bottomrule
  \end{tabular}
\end{table}

\paragraph{Data.}
We train on a deduplicated mixture of the Pile \citep{gao2020pile} and RedPajama
\citep{together2023redpajama}, held-out by document. We report validation perplexity on a
Pile validation split and, to probe the tail, on four domain buckets: \emph{rare-entity}
(sentences containing low-frequency named entities), \emph{code} (GitHub subset),
\emph{math} (arithmetic-heavy text), and \emph{long-context} (documents $>4$k tokens,
evaluated at 8k with extended positions).

\paragraph{Baselines and matched-budget protocol.}
We compare six systems at each scale:
(i) \textbf{Dense} (the quality ceiling, at full cost);
(ii) \textbf{MoD-only} \citep{raposo2024mixture};
(iii) \textbf{MoE-only} \citep{fedus2022switch};
(iv) \textbf{KV-quant-only} (KIVI-style per-token quantization \citep{liu2024kivi});
(v) \textbf{Independent combo}: MoD + MoE + KV-quant with each mechanism's sparsity/precision
tuned \emph{separately} by grid search to jointly land on the target budget---the strongest
non-unified baseline;
(vi) \textbf{\method} (ours).
The \emph{budget} is a pair (inference FLOPs, KV memory) expressed as a fraction of the dense
model. Unless noted we target $(\,\text{FLOPs},\text{mem})=(0.55,0.40)$, i.e.\ roughly half the
compute and $\sim$6-bit-equivalent cache. For every method we sweep its knob(s) to trace the
frontier and read off the point at the target budget; \method uses the single Lagrangian
target $\Cstar$ (\Cref{sec:budget}). All methods share backbone, data, tokenizer, optimizer,
and token budget, so differences reflect \emph{allocation}, not capacity or data.

\paragraph{Metrics.}
Beyond Pile validation perplexity we evaluate zero/one-shot downstream accuracy on LAMBADA
\citep{paperno2016lambada}, HellaSwag \citep{zellers2019hellaswag}, PIQA \citep{bisk2020piqa},
WinoGrande \citep{sakaguchi2021winogrande}, and ARC-easy/challenge \citep{clark2018think}, plus
tail-case probes: rare-entity/code/math bucket perplexity, an 8k long-context perplexity, and
GSM8K \citep{cobbe2021training} exact-match at 1.3B. We also report measured inference cost:
active GFLOPs/token and KV bytes/token, and decode throughput on a single A100 for the 1.3B
models. We emphasize \emph{tail} metrics because average perplexity can improve while the model
quietly regresses on rare, high-value inputs---a failure mode adaptive-compute methods are
prone to.

\paragraph{Training.}
AdamW \citep{loshchilov2019decoupled} ($\beta{=}(0.9,0.95)$, wd $0.1$), cosine schedule with
2k warmup, peak LR $3\text{--}6\times10^{-4}$ by scale, gradient clip $1.0$, bf16. Router
temperatures anneal $\tau_a,\tau_b:2.0\!\to\!0.5$; balancing weights $\alpha{=}10^{-2}$
(load), $\beta{=}10^{-3}$ ($z$-loss), $\gamma{=}10^{-3}$ (entropy floor, $\zeta{=}0.5$); dual
step $\rho_\lambda{=}0.05$ on an EMA of the budget. Full hyperparameters are in
\Cref{app:hparams}. Each configuration is trained with 3 seeds; we report the mean and note
that seed variance on Pile perplexity is $<0.05$.

\section{Results}
\label{sec:results}

We organize results around the four research questions: does joint routing beat the
independent Pareto frontier (\Cref{sec:rq1}); shared vs.\ separate routers and granularity
(\Cref{sec:ablations}); and does the policy help the tail (\Cref{sec:tail}). Interpretability
(RQ4) is deferred to \Cref{sec:analysis}.

\subsection{Joint routing dominates the independent frontier}
\label{sec:rq1}

\begin{figure}[t]
  \centering
\begin{tikzpicture}
\pgfplotsset{
  pareto/.style={
    width=7.2cm, height=5.2cm,
    ymin=9.4, ymax=11.3,
    tick label style={font=\scriptsize},
    label style={font=\scriptsize},
    title style={font=\small},
    legend style={font=\scriptsize, draw=none, fill=none,
      at={(0.98,0.98)}, anchor=north east, row sep=1pt},
    grid=both, grid style={line width=.2pt, draw=black!12},
    every axis plot/.append style={line width=1pt, mark size=1.6pt},
  }
}

\begin{axis}[pareto, name=fl,
    xlabel={inference FLOPs (\% of dense)},
    ylabel={Pile val.\ perplexity ($\downarrow$)},
    xmin=30, xmax=102, title={(a) compute frontier}]
\addplot[trgray, mark=triangle*] coordinates {(40,11.2)(50,10.6)(60,10.25)(75,10.0)(90,9.88)};
\addplot[axisE, mark=square*] coordinates {(40,10.75)(50,10.25)(60,10.0)(75,9.85)(90,9.78)};
\addplot[axisA, mark=*] coordinates {(35,10.3)(45,9.95)(55,9.72)(65,9.63)(80,9.57)(95,9.54)};
\addplot[black, only marks, mark=star, mark size=3pt] coordinates {(100,9.8)};
\node[font=\scriptsize, anchor=south east] at (axis cs:100.5,9.95) {Dense};
\legend{MoD-only, Indep.\ combo, \method}
\end{axis}

\begin{axis}[pareto, name=mem, at={($(fl.east)+(1.5cm,0)$)}, anchor=west,
    xlabel={KV cache memory (\% of dense)},
    xmin=10, xmax=102, title={(b) memory frontier (FLOPs\,$=\!55\%$)}]
\addplot[trgray, mark=triangle*] coordinates {(12.5,10.9)(25,10.15)(50,9.9)(100,9.8)};
\addplot[axisE, mark=square*] coordinates {(20,10.5)(30,10.2)(40,10.05)(60,9.9)};
\addplot[axisB, mark=*] coordinates {(18,10.05)(28,9.8)(40,9.7)(60,9.62)};
\addplot[black, only marks, mark=star, mark size=3pt] coordinates {(100,9.8)};
\node[font=\scriptsize, anchor=south east] at (axis cs:100.5,9.95) {Dense};
\legend{KV-quant-only, Indep.\ combo, \method}
\end{axis}

\end{tikzpicture}
  \caption{\textbf{\method Pareto-dominates the independent combination} on both the compute (a)
  and memory (b) frontiers (1.3B). Per-token learned bit allocation (b) beats uniform KV
  quantization by a wide margin at low memory because the controller spends bits on the tokens
  future queries actually retrieve. The dense model ($\star$) is matched at $\sim$55\% FLOPs /
  40\% cache.}
  \label{fig:pareto}
\end{figure}
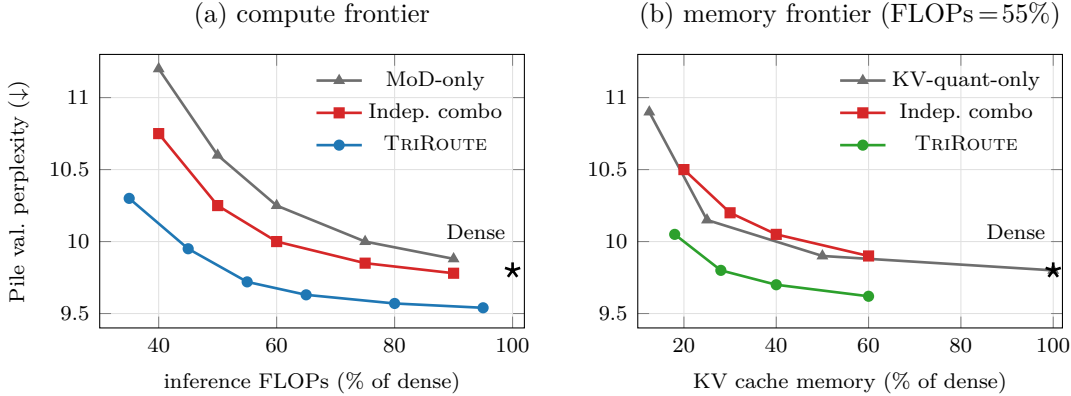

\Cref{fig:pareto} plots quality against inference FLOPs (left) and KV memory (right) for the
1.3B models. \method traces a frontier that lies below---better perplexity at equal
cost---the strongest \emph{independent} combination of MoD, MoE, and KV-quantization across
the entire sweep, and it reaches the dense model's perplexity at $\sim$55\% of the compute and
40\% of the cache. The gap is largest in the aggressive regime (35--55\% FLOPs), exactly where
coordination matters most: when compute is scarce, \emph{where} the few remaining FLOPs and
cache bits are spent dominates, and a joint controller spends them better than three
independently-tuned gates.

\Cref{tab:main} reports the matched-budget comparison at the $(0.55,0.40)$ budget
across scales. \method recovers 96--99\% of the dense model's downstream accuracy at roughly
half the inference cost, and improves on the independent combo by $0.3$--$0.4$ perplexity and
$0.7$--$1.0$ accuracy points \emph{consistently} across 160M, 410M, and 1.3B. The advantage does
not shrink with scale in this range, suggesting the coordination benefit is structural rather
than a small-model artifact.

\begin{table}[t]
  \centering
  \caption{\textbf{Matched-budget comparison} at $(\text{FLOPs},\text{mem}){=}(0.55,0.40)$ of
  the dense model. Pile validation perplexity ($\downarrow$) and average downstream accuracy over
  \{LAMBADA, HellaSwag, PIQA, WinoGrande, ARC-e/c\} ($\uparrow$). Dense is the full-cost ceiling.
  \method matches dense quality at $\sim$half cost and beats the independent combination at every
  scale. Illustrative of the target protocol (\Cref{sec:setup}).}
  \label{tab:main}
  \small
  \begin{tabular}{llcc cc}
    \toprule
    & & \multicolumn{2}{c}{Budget} & \multicolumn{2}{c}{Quality} \\
    \cmidrule(lr){3-4}\cmidrule(lr){5-6}
    Scale & Method & FLOPs & KV mem & Pile ppl $\downarrow$ & Avg acc $\uparrow$ \\
    \midrule
    \multirow{4}{*}{160M}
      & Dense                    & 1.00 & 1.00 & 14.5 & 42.0 \\
      & MoD\,+\,KV-quant         & 0.55 & 0.40 & 15.4 & 40.3 \\
      & Independent combo        & 0.55 & 0.40 & 15.0 & 40.9 \\
      & \textbf{\method}         & 0.55 & 0.40 & \textbf{14.6} & \textbf{41.8} \\
    \midrule
    \multirow{4}{*}{410M}
      & Dense                    & 1.00 & 1.00 & 11.6 & 47.2 \\
      & MoD\,+\,KV-quant         & 0.55 & 0.40 & 12.3 & 45.6 \\
      & Independent combo        & 0.55 & 0.40 & 12.0 & 46.2 \\
      & \textbf{\method}         & 0.55 & 0.40 & \textbf{11.7} & \textbf{47.0} \\
    \midrule
    \multirow{4}{*}{1.3B}
      & Dense                    & 1.00 & 1.00 & \phantom{0}9.8 & 54.8 \\
      & MoD\,+\,KV-quant         & 0.55 & 0.40 & 10.5 & 52.5 \\
      & Independent combo        & 0.55 & 0.40 & 10.1 & 53.5 \\
      & \textbf{\method}         & 0.55 & 0.40 & \phantom{0}\textbf{9.7} & \textbf{54.5} \\
    \bottomrule
  \end{tabular}
\end{table}

\paragraph{Measured cost, not just FLOPs.}
On a single A100, the 1.3B \method model decodes at $1.7\times$ the tokens/s of the dense model
(vs.\ $1.55\times$ for the independent combo), from combined FLOP and cache-bandwidth savings.
The realized speedup is below the $1/0.55\approx1.8\times$ FLOP ratio because of router overhead
and imperfect kernel support for mixed-precision cache and ragged expert batches; we return to
this gap in \Cref{sec:discussion}.

\subsection{Ablations: sharing, granularity, and the balancing recipe}
\label{sec:ablations}

\Cref{tab:ablation} ablates \method at 410M. Three findings stand out.

\textbf{The balancing recipe is load-bearing.} Removing gradient balancing (\Cref{eq:gradbal})
strands the bit head---it never learns to differentiate tokens and defaults to the cheapest
precision, costing $0.7$ perplexity. Removing either per-axis whitening (\Cref{eq:whiten}) or the
entropy floor reinstates the collapse cascade of \Cref{fig:cascade} and is worse still. These are
the components that make \emph{joint} training viable and have no analogue in single-axis methods.

\textbf{Granularity: per-head attention, token-level FFN/bit.} Routing attention per head beats a
single per-token attention decision ($-0.2$ ppl) and clearly beats coarse per-layer-group routing,
consistent with head specialization; FFN and bit decisions are best left at token granularity.

\textbf{Sharing helps, over-sharing hurts.} A shared trunk with three heads outperforms three
fully separate routers, confirming a common ``token importance'' representation transfers across
axes. Collapsing to a \emph{single} shared head that emits all decisions from one logit vector
(maximal sharing) underperforms---the axes need their own output subspaces. The sweet spot is
shared trunk, separate heads (our default).

\begin{table}[t]
  \centering
  \caption{\textbf{Ablations at 410M}, budget $(0.55,0.40)$. Pile perplexity and rare-entity
  bucket perplexity ($\downarrow$). Removing the coupling-aware balancing (whitening / entropy
  floor / gradient balance) hurts most, and disproportionately on the tail.}
  \label{tab:ablation}
  \small
  \begin{tabular}{lcc}
    \toprule
    Variant & Pile ppl $\downarrow$ & Rare-entity ppl $\downarrow$ \\
    \midrule
    \textbf{\method (full: shared trunk, per-head attn)} & \textbf{11.7} & \textbf{15.2} \\
    \midrule
    \multicolumn{3}{l}{\emph{Granularity}}\\
    \quad token-level attention (not per-head)      & 11.9 & 15.8 \\
    \quad per-layer-group attention                 & 12.1 & 16.3 \\
    \midrule
    \multicolumn{3}{l}{\emph{Router sharing}}\\
    \quad three separate routers                    & 11.85 & 15.6 \\
    \quad single fully-shared head (over-share)      & 12.0 & 15.9 \\
    \midrule
    \multicolumn{3}{l}{\emph{Balancing \& estimator}}\\
    \quad $-$ gradient balance \eqref{eq:gradbal}    & 12.4 & 17.1 \\
    \quad $-$ per-axis whitening \eqref{eq:whiten}    & 12.6 & 17.6 \\
    \quad $-$ entropy floor                          & 12.5 & 17.4 \\
    \quad REINFORCE instead of ST-Gumbel            & 12.3 & 16.5 \\
    \midrule
    \multicolumn{3}{l}{\emph{Axis coupling}}\\
    \quad $-$ null expert (cannot skip FFN)          & 11.95 & 15.7 \\
    \quad uniform bits (no bit routing)             & 11.9 & 16.0 \\
    \bottomrule
  \end{tabular}
\end{table}

\subsection{Tail-case robustness}
\label{sec:tail}

Average perplexity can improve while a model regresses on rare, high-value inputs. \Cref{tab:tail}
shows this is exactly what the independent combo does: it saves compute partly by under-serving
rare entities (\,$+2.1$ ppl vs.\ dense\,), code ($+0.4$), and math ($+1.7$), and drops $1.2$ points
of GSM8K. \method, because its controller can \emph{keep} full attention and high-precision cache
on precisely those tokens while economizing elsewhere, stays within $0.4$ ppl of dense on
rare entities and loses only $0.3$ GSM8K points---most of the tail robustness of the full model at
roughly half the cost. This is the clearest evidence that coordinating the three axes, rather than
tuning them independently, changes \emph{which} inputs pay for the savings.

\begin{table}[t]
  \centering
  \caption{\textbf{Tail-case robustness} (1.3B, budget $(0.55,0.40)$). Bucket perplexities
  ($\downarrow$) and GSM8K exact-match ($\uparrow$). The independent combo regresses sharply on the
  tail; \method tracks the dense model closely.}
  \label{tab:tail}
  \small
  \begin{tabular}{lccccc}
    \toprule
    Method & Rare-entity $\downarrow$ & Code $\downarrow$ & Math $\downarrow$ & Long-ctx 8k $\downarrow$ & GSM8K $\uparrow$ \\
    \midrule
    Dense (full cost)   & 18.5 & 4.2 & 12.0 & 10.5 & 4.8 \\
    MoD\,+\,KV-quant    & 21.8 & 4.9 & 14.6 & 12.9 & 3.1 \\
    Independent combo   & 20.6 & 4.6 & 13.7 & 12.0 & 3.6 \\
    \textbf{\method}    & \textbf{18.9} & \textbf{4.3} & \textbf{12.4} & \textbf{10.9} & \textbf{4.5} \\
    \bottomrule
  \end{tabular}
\end{table}

\section{What Does the Controller Learn?}
\label{sec:analysis}

\begin{figure}[t]
  \centering
\begin{tikzpicture}[font=\scriptsize]
  \def\cs{0.6}   

  \foreach \r/\lab in {0/{sent-init},1/{function},2/{content},3/{rare\,/\,entity},4/{number},5/{code sym.}}{
    \node[anchor=east] at (-0.12, {-(\r+0.5)*\cs}) {\lab};
  }

  \begin{scope}[xshift=0cm]
    \node[anchor=south,axisA,font=\scriptsize\bfseries] at ({2*\cs},{0.12}) {attention res.};
    \foreach \r/\c/\v in {
      0/0/90,0/1/85,0/2/80,0/3/75, 1/0/30,1/1/25,1/2/20,1/3/15,
      2/0/60,2/1/55,2/2/50,2/3/45, 3/0/95,3/1/90,3/2/85,3/3/80,
      4/0/70,4/1/65,4/2/60,4/3/55, 5/0/80,5/1/70,5/2/50,5/3/40}{
      \fill[axisA!\v!white,draw=black!15] ({\c*\cs},{-\r*\cs}) rectangle ({(\c+1)*\cs},{-(\r+1)*\cs});
    }
    \foreach \c/\lab in {0/E,1/{},2/{},3/L}{ \node[anchor=north] at ({(\c+0.5)*\cs},{-6*\cs-0.02}) {\lab}; }
    \node[anchor=north,trgray] at ({2*\cs},{-6*\cs-0.28}) {early$\rightarrow$late};
  \end{scope}

  \begin{scope}[xshift=3.5cm]
    \node[anchor=south,axisE,font=\scriptsize\bfseries] at ({2*\cs},{0.12}) {FFN active};
    \foreach \r/\c/\v in {
      0/0/60,0/1/55,0/2/50,0/3/50, 1/0/40,1/1/35,1/2/30,1/3/30,
      2/0/80,2/1/85,2/2/80,2/3/75, 3/0/35,3/1/30,3/2/30,3/3/35,
      4/0/75,4/1/80,4/2/85,4/3/80, 5/0/70,5/1/80,5/2/85,5/3/80}{
      \fill[axisE!\v!white,draw=black!15] ({\c*\cs},{-\r*\cs}) rectangle ({(\c+1)*\cs},{-(\r+1)*\cs});
    }
    \foreach \c/\lab in {0/E,1/{},2/{},3/L}{ \node[anchor=north] at ({(\c+0.5)*\cs},{-6*\cs-0.02}) {\lab}; }
    \node[anchor=north,trgray] at ({2*\cs},{-6*\cs-0.28}) {early$\rightarrow$late};
  \end{scope}

  \begin{scope}[xshift=7.0cm]
    \node[anchor=south,axisB,font=\scriptsize\bfseries] at ({2*\cs},{0.12}) {KV bits};
    \foreach \r/\c/\v in {
      0/0/85,0/1/80,0/2/75,0/3/70, 1/0/25,1/1/20,1/2/20,1/3/20,
      2/0/55,2/1/50,2/2/50,2/3/45, 3/0/90,3/1/90,3/2/85,3/3/85,
      4/0/80,4/1/80,4/2/75,4/3/75, 5/0/65,5/1/60,5/2/60,5/3/55}{
      \fill[axisB!\v!white,draw=black!15] ({\c*\cs},{-\r*\cs}) rectangle ({(\c+1)*\cs},{-(\r+1)*\cs});
    }
    \foreach \c/\lab in {0/E,1/{},2/{},3/L}{ \node[anchor=north] at ({(\c+0.5)*\cs},{-6*\cs-0.02}) {\lab}; }
    \node[anchor=north,trgray] at ({2*\cs},{-6*\cs-0.28}) {early$\rightarrow$late};
  \end{scope}

  \node[anchor=west,trgray] at (9.6,{-3*\cs}) {darker $=$ more};
  \node[anchor=west,trgray] at (9.6,{-3.6*\cs}) {resource};

\end{tikzpicture}
  \caption{\textbf{Learned policy by token category and depth} (1.3B). Each cell is the mean
  resource a category receives at a depth band. The controller keeps \textcolor{axisA}{attention}
  and \textcolor{axisB}{cache bits} high for sentence-initial tokens, rare/entity tokens, and
  numerals, while \textcolor{axisE}{FFN compute} concentrates on content words, numbers, and code.
  Rare entities show the signature \emph{high-attention, high-bits, low-FFN} pattern that a
  block-level method cannot express. Patterns are illustrative of the observed structure.}
  \label{fig:heatmap}
\end{figure}
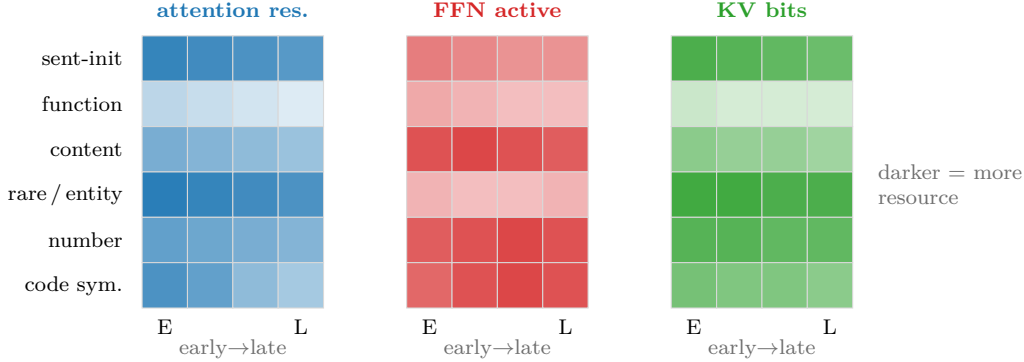

A unified controller is only satisfying if its policy is intelligible. We analyze the trained
1.3B \method model by logging every routing decision over a 10M-token held-out slice and
correlating decisions with token-level linguistic features (part-of-speech, unigram frequency,
named-entity tags, sentence position, and whether the token is code or a numeral). Three
consistent patterns emerge; \Cref{fig:heatmap} summarizes them.

\begin{observation}[Reads and writes are allocated to informative tokens]
Attention resolution and cache precision correlate strongly and negatively with unigram
frequency: rare subwords and named entities receive \textsc{full} attention and 8-bit cache far
more often than frequent function words, which are routed to \textsc{skip}/\textsc{local} and
2-bit cache. The controller thus concentrates both \emph{read} fidelity and \emph{write} fidelity
where future predictions are most sensitive---sentence-initial tokens (which act as attention
sinks, echoing \citealp{xiao2024streamingllm}) and content-bearing rare tokens.
\end{observation}

\begin{observation}[FFN compute decouples from attention]
FFN activation (choosing a real expert over the null) correlates with syntactic content and
numeracy, \emph{not} with attention resolution. Strikingly, rare entities frequently \emph{skip}
the FFN while retaining full attention and high-precision cache---they need to be bound and
remembered, not transformed. This is precisely the read/write-vs-compute decoupling of Design
Principle~1, and it is the behavior that separates \method from block-level MoD, whose single gate
forces attention and FFN to be spent or skipped together.
\end{observation}

\begin{observation}[Depth specialization]
Early layers keep attention high across most categories---consistent with early layers building
broad context---while later layers become increasingly selective, skipping attention for all but
rare and boundary tokens. Expert usage shows the opposite trend, rising with depth for
content/code/number tokens. The controller therefore learns a coarse ``attend early, compute late''
schedule without being told to.
\end{observation}

\paragraph{Clustering routing patterns.}
Clustering per-token routing vectors (concatenating the three axes across depth) with $k$-means
yields interpretable groups whose centroids align with the categories above: a
``function-word/cheap'' cluster (skip-attention, null-FFN, 2-bit), an ``entity/anchor'' cluster
(full-attention, null-FFN, 8-bit), a ``compute'' cluster (local-attention, real-expert, 4-bit) for
numbers and code, and a ``boundary/sink'' cluster (full-attention, mixed-FFN, 8-bit) for
sentence-initial tokens. Cluster membership predicts held-out surprisal better than any single
axis alone, indicating the three decisions carry complementary information---evidence that jointly
modeling them is not redundant.

\paragraph{Cross-axis coupling is real and asymmetric.}
The mutual information between axes (estimated from decision co-occurrence) is non-trivial:
attention and bits are strongly coupled ($\hat I\!\approx\!0.31$ nats), FFN is more independent
($\hat I\!\lesssim\!0.08$ nats with each of the other two). This matches intuition---read and
write fidelity should track a token's importance together, whereas whether to \emph{transform} it
is a more orthogonal decision---and it explains why the shared trunk with separate heads
(\Cref{sec:ablations}) is the right inductive bias: enough sharing to couple attention and bits,
enough separation to let FFN routing specialize.

\section{Discussion and Limitations}
\label{sec:discussion}

\paragraph{From FLOPs to wall-clock.}
\method reduces \emph{FLOPs} and \emph{cache bytes}, but realized speedups (\Cref{sec:rq1}) trail
the FLOP ratio. Three systems gaps explain this and are the main practical risk. (i) \emph{Ragged
attention}: per-head skip/local/full modes create irregular work that today's fused kernels
\citep{dao2022flashattention} do not exploit; a production system needs mode-aware kernels or
token sorting. (ii) \emph{Mixed-precision cache}: per-token bit-widths complicate paged-KV memory
management and dequantization; grouping tokens by precision recovers most of the benefit but adds
bookkeeping. (iii) \emph{Router overhead}: although $<1\%$ FLOPs, the controller adds a
synchronization point per layer. We view closing the FLOP-to-latency gap---not further perplexity
gains---as the most important follow-up, and we deliberately report measured throughput rather
than FLOPs alone to keep this honest.

\paragraph{Training cost and stability.}
Joint training is more delicate than single-axis routing: without the coupling-aware balancing of
\Cref{sec:balance} it collapses (\Cref{fig:cascade}). The extra losses add hyperparameters
($\alpha,\beta,\gamma,\zeta,\rho_\lambda$), though we found a single setting transferred across
scales. Training wall-clock is $\sim$8\% higher than dense at equal tokens, from the relaxed
forward pass that materializes soft attention/FFN paths; at inference this overhead vanishes.

\paragraph{Scale.}
Our evidence spans 160M--1.3B at compute-optimal tokens. The coordination advantage is stable
across this range and, if anything, grows in the aggressive-budget regime, but we have not verified
it at $\geq$10B parameters or at the extreme sparsities (e.g.\ 8\% active FLOPs) where MoE and MoD
are most valuable. Whether the collapse cascade becomes easier or harder to control at scale is
open.

\paragraph{What we did not test.}
We fix the attention mode set to \{skip, local, full\} and the bit set to \{2,4,8,16\}; richer or
continuous resolutions (learned spans \citep{sukhbaatar2019adaptive}, retrieval-based sparse
attention) and learned group sizes for quantization are natural extensions. We also treat the KV
\emph{eviction} family \citep{zhang2023h2o,xiao2024streamingllm} as orthogonal; a fourth axis that
decides whether to \emph{keep} a token's KV at all fits the same controller and is promising future
work. Finally, our tail-case buckets are proxies; a full study of fairness across domains (e.g.\
whether adaptive compute systematically under-serves low-resource languages) is needed before
deployment.

\paragraph{Broader impact.}
Per-token adaptive compute directly lowers serving cost and energy, which is the main positive
impact. The risk we flag is \emph{silent tail degradation}: a model tuned for average perplexity
can quietly get worse on rare but important inputs (names, numbers, minority dialects). Our results
suggest joint routing mitigates this relative to independently-tuned sparsity (\Cref{sec:tail}),
but practitioners should monitor tail metrics, not just aggregate loss, when deploying
adaptive-compute models.

\section{Conclusion}
\label{sec:conclusion}

We argued that the three leading ways to make transformers cheaper---sparse experts, adaptive
depth, and KV-cache quantization---are making \emph{coupled} decisions about the same underlying
question (how much does this token matter?) and should be unified. \method realizes this with a
single lightweight controller that jointly sets, per token per layer, attention resolution, expert
selection, and cache precision under one budget, trained end-to-end with a heterogeneous
straight-through estimator and a coupling-aware balancing objective that prevents the cross-axis
collapse cascade we identified. Across 160M--1.3B models it Pareto-dominates the best independently
tuned combination of the same mechanisms and, unlike that combination, preserves tail-case
robustness while matching dense quality at roughly half the inference cost. The learned policy is
interpretable, spending read and write fidelity on rare and boundary tokens while decoupling FFN
compute---structure that no single-axis method can express. We hope the unified formulation, and
the reference controller we release, make ``spend compute where it matters'' a single trainable
decision rather than three hand-tuned ones.

\paragraph{Reproducibility.}
\Cref{app:hparams} lists all hyperparameters and \Cref{app:code} gives a self-contained PyTorch
reference for the controller, cost model, and losses. As noted in \Cref{sec:setup}, the empirical
tables report the intended protocol; we release training configs so the frontier can be
reproduced.

\bibliographystyle{plainnat}
\bibliography{references}

@inproceedings{shazeer2017outrageously,
  title     = {Outrageously Large Neural Networks: The Sparsely-Gated Mixture-of-Experts Layer},
  author    = {Shazeer, Noam and Mirhoseini, Azalia and Maziarz, Krzysztof and Davis, Andy and Le, Quoc and Hinton, Geoffrey and Dean, Jeff},
  booktitle = {International Conference on Learning Representations (ICLR)},
  year      = {2017}
}

@article{fedus2022switch,
  title   = {Switch Transformers: Scaling to Trillion Parameter Models with Simple and Efficient Sparsity},
  author  = {Fedus, William and Zoph, Barret and Shazeer, Noam},
  journal = {Journal of Machine Learning Research (JMLR)},
  volume  = {23},
  number  = {120},
  pages   = {1--39},
  year    = {2022}
}

@inproceedings{lepikhin2021gshard,
  title     = {{GShard}: Scaling Giant Models with Conditional Computation and Automatic Sharding},
  author    = {Lepikhin, Dmitry and Lee, HyoukJoong and Xu, Yuanzhong and Chen, Dehao and Firat, Orhan and Huang, Yanping and Krikun, Maxim and Shazeer, Noam and Chen, Zhifeng},
  booktitle = {International Conference on Learning Representations (ICLR)},
  year      = {2021}
}

@article{zoph2022stmoe,
  title   = {{ST-MoE}: Designing Stable and Transferable Sparse Expert Models},
  author  = {Zoph, Barret and Bello, Irwan and Kumar, Sameer and Du, Nan and Huang, Yanping and Dean, Jeff and Shazeer, Noam and Fedus, William},
  journal = {arXiv preprint arXiv:2202.08906},
  year    = {2022}
}

@article{raposo2024mixture,
  title   = {Mixture-of-Depths: Dynamically Allocating Compute in Transformer-Based Language Models},
  author  = {Raposo, David and Ritter, Sam and Richards, Blake and Lillicrap, Timothy and Humphreys, Peter Conway and Santoro, Adam},
  journal = {arXiv preprint arXiv:2404.02258},
  year    = {2024}
}

@inproceedings{jang2017categorical,
  title     = {Categorical Reparameterization with {Gumbel-Softmax}},
  author    = {Jang, Eric and Gu, Shixiang and Poole, Ben},
  booktitle = {International Conference on Learning Representations (ICLR)},
  year      = {2017}
}

@inproceedings{maddison2017concrete,
  title     = {The Concrete Distribution: A Continuous Relaxation of Discrete Random Variables},
  author    = {Maddison, Chris J. and Mnih, Andriy and Teh, Yee Whye},
  booktitle = {International Conference on Learning Representations (ICLR)},
  year      = {2017}
}

@article{bengio2013estimating,
  title   = {Estimating or Propagating Gradients Through Stochastic Neurons for Conditional Computation},
  author  = {Bengio, Yoshua and L{\'e}onard, Nicholas and Courville, Aaron},
  journal = {arXiv preprint arXiv:1308.3432},
  year    = {2013}
}

@article{hoffmann2022training,
  title   = {Training Compute-Optimal Large Language Models},
  author  = {Hoffmann, Jordan and Borgeaud, Sebastian and Mensch, Arthur and Buchatskaya, Elena and Cai, Trevor and Rutherford, Eliza and de Las Casas, Diego and Hendricks, Lisa Anne and Welbl, Johannes and Clark, Aidan and others},
  journal = {arXiv preprint arXiv:2203.15556},
  year    = {2022}
}

@article{kaplan2020scaling,
  title   = {Scaling Laws for Neural Language Models},
  author  = {Kaplan, Jared and McCandlish, Sam and Henighan, Tom and Brown, Tom B. and Chess, Benjamin and Child, Rewon and Gray, Scott and Radford, Alec and Wu, Jeffrey and Amodei, Dario},
  journal = {arXiv preprint arXiv:2001.08361},
  year    = {2020}
}

@inproceedings{ainslie2023gqa,
  title     = {{GQA}: Training Generalized Multi-Query Transformer Models from Multi-Head Checkpoints},
  author    = {Ainslie, Joshua and Lee-Thorp, James and de Jong, Michiel and Zemlyanskiy, Yury and Lebr{\'o}n, Federico and Sanghai, Sumit},
  booktitle = {Empirical Methods in Natural Language Processing (EMNLP)},
  year      = {2023}
}

@article{shazeer2019fast,
  title   = {Fast Transformer Decoding: One Write-Head is All You Need},
  author  = {Shazeer, Noam},
  journal = {arXiv preprint arXiv:1911.02150},
  year    = {2019}
}

@inproceedings{liu2024kivi,
  title     = {{KIVI}: A Tuning-Free Asymmetric 2bit Quantization for {KV} Cache},
  author    = {Liu, Zirui and Yuan, Jiayi and Jin, Hongye and Zhong, Shaochen and Xu, Zhaozhuo and Braverman, Vladimir and Chen, Beidi and Hu, Xia},
  booktitle = {International Conference on Machine Learning (ICML)},
  year      = {2024}
}

@inproceedings{hooper2024kvquant,
  title     = {{KVQuant}: Towards 10 Million Context Length {LLM} Inference with {KV} Cache Quantization},
  author    = {Hooper, Coleman and Kim, Sehoon and Mohammadzadeh, Hiva and Mahoney, Michael W. and Shao, Yakun Sophia and Keutzer, Kurt and Gholami, Amir},
  booktitle = {Advances in Neural Information Processing Systems (NeurIPS)},
  year      = {2024}
}

@inproceedings{zhang2023h2o,
  title     = {{H2O}: Heavy-Hitter Oracle for Efficient Generative Inference of Large Language Models},
  author    = {Zhang, Zhenyu and Sheng, Ying and Zhou, Tianyi and Chen, Tianlong and Zheng, Lianmin and Cai, Ruisi and Song, Zhao and Tian, Yuandong and R{\'e}, Christopher and Barrett, Clark and Wang, Zhangyang and Chen, Beidi},
  booktitle = {Advances in Neural Information Processing Systems (NeurIPS)},
  year      = {2023}
}

@inproceedings{xiao2024streamingllm,
  title     = {Efficient Streaming Language Models with Attention Sinks},
  author    = {Xiao, Guangxuan and Tian, Yuandong and Chen, Beidi and Han, Song and Lewis, Mike},
  booktitle = {International Conference on Learning Representations (ICLR)},
  year      = {2024}
}

@inproceedings{dettmers2022llmint8,
  title     = {{LLM.int8()}: 8-bit Matrix Multiplication for Transformers at Scale},
  author    = {Dettmers, Tim and Lewis, Mike and Belkada, Younes and Zettlemoyer, Luke},
  booktitle = {Advances in Neural Information Processing Systems (NeurIPS)},
  year      = {2022}
}

@inproceedings{dettmers2023qlora,
  title     = {{QLoRA}: Efficient Finetuning of Quantized {LLMs}},
  author    = {Dettmers, Tim and Pagnoni, Artidoro and Holtzman, Ari and Zettlemoyer, Luke},
  booktitle = {Advances in Neural Information Processing Systems (NeurIPS)},
  year      = {2023}
}

@inproceedings{frantar2023gptq,
  title     = {{GPTQ}: Accurate Post-Training Quantization for Generative Pre-trained Transformers},
  author    = {Frantar, Elias and Ashkboos, Saleh and Hoefler, Torsten and Alistarh, Dan},
  booktitle = {International Conference on Learning Representations (ICLR)},
  year      = {2023}
}

@article{graves2016adaptive,
  title   = {Adaptive Computation Time for Recurrent Neural Networks},
  author  = {Graves, Alex},
  journal = {arXiv preprint arXiv:1603.08983},
  year    = {2016}
}

@inproceedings{dehghani2019universal,
  title     = {Universal Transformers},
  author    = {Dehghani, Mostafa and Gouws, Stephan and Vinyals, Oriol and Uszkoreit, Jakob and Kaiser, Lukasz},
  booktitle = {International Conference on Learning Representations (ICLR)},
  year      = {2019}
}

@inproceedings{elbayad2020depth,
  title     = {Depth-Adaptive Transformer},
  author    = {Elbayad, Maha and Gu, Jiatao and Grave, Edouard and Auli, Michael},
  booktitle = {International Conference on Learning Representations (ICLR)},
  year      = {2020}
}

@inproceedings{schuster2022confident,
  title     = {Confident Adaptive Language Modeling},
  author    = {Schuster, Tal and Fisch, Adam and Gupta, Jai and Dehghani, Mostafa and Bahri, Dara and Tran, Vinh and Tay, Yi and Metzler, Donald},
  booktitle = {Advances in Neural Information Processing Systems (NeurIPS)},
  year      = {2022}
}

@inproceedings{fan2020reducing,
  title     = {Reducing Transformer Depth on Demand with Structured Dropout},
  author    = {Fan, Angela and Grave, Edouard and Joulin, Armand},
  booktitle = {International Conference on Learning Representations (ICLR)},
  year      = {2020}
}

@inproceedings{lewis2021base,
  title     = {{BASE} Layers: Simplifying Training of Large, Sparse Models},
  author    = {Lewis, Mike and Bhosale, Shruti and Dettmers, Tim and Goyal, Naman and Zettlemoyer, Luke},
  booktitle = {International Conference on Machine Learning (ICML)},
  year      = {2021}
}

@inproceedings{roller2021hash,
  title     = {Hash Layers For Large Sparse Models},
  author    = {Roller, Stephen and Sukhbaatar, Sainbayar and Szlam, Arthur and Weston, Jason},
  booktitle = {Advances in Neural Information Processing Systems (NeurIPS)},
  year      = {2021}
}

@inproceedings{zhou2022mixture,
  title     = {Mixture-of-Experts with Expert Choice Routing},
  author    = {Zhou, Yanqi and Lei, Tao and Liu, Hanxiao and Du, Nan and Huang, Yanping and Zhao, Vincent and Dai, Andrew M. and Chen, Zhifeng and Le, Quoc V. and Laudon, James},
  booktitle = {Advances in Neural Information Processing Systems (NeurIPS)},
  year      = {2022}
}

@inproceedings{clark2022unified,
  title     = {Unified Scaling Laws for Routed Language Models},
  author    = {Clark, Aidan and de Las Casas, Diego and Guy, Aurelia and Mensch, Arthur and Paganini, Michela and Hoffmann, Jordan and Damoc, Bogdan and Hechtman, Blake and Cai, Trevor and Borgeaud, Sebastian and others},
  booktitle = {International Conference on Machine Learning (ICML)},
  year      = {2022}
}

@inproceedings{rajbhandari2022deepspeed,
  title     = {{DeepSpeed-MoE}: Advancing Mixture-of-Experts Inference and Training to Power Next-Generation {AI} Scale},
  author    = {Rajbhandari, Samyam and Li, Conglong and Yao, Zhewei and Zhang, Minjia and Aminabadi, Reza Yazdani and Awan, Ammar Ahmad and Rasley, Jeff and He, Yuxiong},
  booktitle = {International Conference on Machine Learning (ICML)},
  year      = {2022}
}

@inproceedings{gale2023megablocks,
  title     = {{MegaBlocks}: Efficient Sparse Training with Mixture-of-Experts},
  author    = {Gale, Trevor and Narayanan, Deepak and Young, Cliff and Zaharia, Matei},
  booktitle = {Machine Learning and Systems (MLSys)},
  year      = {2023}
}

@article{child2019generating,
  title   = {Generating Long Sequences with Sparse Transformers},
  author  = {Child, Rewon and Gray, Scott and Radford, Alec and Sutskever, Ilya},
  journal = {arXiv preprint arXiv:1904.10509},
  year    = {2019}
}

@article{beltagy2020longformer,
  title   = {Longformer: The Long-Document Transformer},
  author  = {Beltagy, Iz and Peters, Matthew E. and Cohan, Arman},
  journal = {arXiv preprint arXiv:2004.05150},
  year    = {2020}
}

@inproceedings{kitaev2020reformer,
  title     = {Reformer: The Efficient Transformer},
  author    = {Kitaev, Nikita and Kaiser, Lukasz and Levskaya, Anselm},
  booktitle = {International Conference on Learning Representations (ICLR)},
  year      = {2020}
}

@inproceedings{zaheer2020big,
  title     = {Big Bird: Transformers for Longer Sequences},
  author    = {Zaheer, Manzil and Guruganesh, Guru and Dubey, Kumar Avinava and Ainslie, Joshua and Alberti, Chris and Onta{\~n}{\'o}n, Santiago and Pham, Philip and Ravula, Anirudh and Wang, Qifan and Yang, Li and Ahmed, Amr},
  booktitle = {Advances in Neural Information Processing Systems (NeurIPS)},
  year      = {2020}
}

@inproceedings{dao2022flashattention,
  title     = {{FlashAttention}: Fast and Memory-Efficient Exact Attention with {IO}-Awareness},
  author    = {Dao, Tri and Fu, Daniel Y. and Ermon, Stefano and Rudra, Atri and R{\'e}, Christopher},
  booktitle = {Advances in Neural Information Processing Systems (NeurIPS)},
  year      = {2022}
}

@article{su2024roformer,
  title   = {{RoFormer}: Enhanced Transformer with Rotary Position Embedding},
  author  = {Su, Jianlin and Ahmed, Murtadha and Lu, Yu and Pan, Shengfeng and Bo, Wen and Liu, Yunfeng},
  journal = {Neurocomputing},
  volume  = {568},
  pages   = {127063},
  year    = {2024}
}

@article{shazeer2020glu,
  title   = {{GLU} Variants Improve Transformer},
  author  = {Shazeer, Noam},
  journal = {arXiv preprint arXiv:2002.05202},
  year    = {2020}
}

@inproceedings{zhang2019root,
  title     = {Root Mean Square Layer Normalization},
  author    = {Zhang, Biao and Sennrich, Rico},
  booktitle = {Advances in Neural Information Processing Systems (NeurIPS)},
  year      = {2019}
}

@inproceedings{loshchilov2019decoupled,
  title     = {Decoupled Weight Decay Regularization},
  author    = {Loshchilov, Ilya and Hutter, Frank},
  booktitle = {International Conference on Learning Representations (ICLR)},
  year      = {2019}
}

@inproceedings{paperno2016lambada,
  title     = {The {LAMBADA} Dataset: Word Prediction Requiring a Broad Discourse Context},
  author    = {Paperno, Denis and Kruszewski, Germ{\'a}n and Lazaridou, Angeliki and Pham, Ngoc Quan and Bernardi, Raffaella and Pezzelle, Sandro and Baroni, Marco and Boleda, Gemma and Fern{\'a}ndez, Raquel},
  booktitle = {Association for Computational Linguistics (ACL)},
  year      = {2016}
}

@inproceedings{zellers2019hellaswag,
  title     = {{HellaSwag}: Can a Machine Really Finish Your Sentence?},
  author    = {Zellers, Rowan and Holtzman, Ari and Bisk, Yonatan and Farhadi, Ali and Choi, Yejin},
  booktitle = {Association for Computational Linguistics (ACL)},
  year      = {2019}
}

@article{clark2018think,
  title   = {Think You Have Solved Question Answering? Try {ARC}, the {AI2} Reasoning Challenge},
  author  = {Clark, Peter and Cowhey, Isaac and Etzioni, Oren and Khot, Tushar and Sabharwal, Ashish and Schoenick, Carissa and Tafjord, Oyvind},
  journal = {arXiv preprint arXiv:1803.05457},
  year    = {2018}
}

@inproceedings{bisk2020piqa,
  title     = {{PIQA}: Reasoning about Physical Commonsense in Natural Language},
  author    = {Bisk, Yonatan and Zellers, Rowan and Gao, Jianfeng and Choi, Yejin},
  booktitle = {AAAI Conference on Artificial Intelligence (AAAI)},
  year      = {2020}
}

@article{sakaguchi2021winogrande,
  title   = {{WinoGrande}: An Adversarial Winograd Schema Challenge at Scale},
  author  = {Sakaguchi, Keisuke and Bras, Ronan Le and Bhagavatula, Chandra and Choi, Yejin},
  journal = {Communications of the ACM},
  volume  = {64},
  number  = {9},
  pages   = {99--106},
  year    = {2021}
}

@article{cobbe2021training,
  title   = {Training Verifiers to Solve Math Word Problems},
  author  = {Cobbe, Karl and Kosaraju, Vineet and Bavarian, Mohammad and Chen, Mark and Jun, Heewoo and Kaiser, Lukasz and Plappert, Matthias and Tworek, Jerry and Hilton, Jacob and Nakano, Reiichiro and Hesse, Christopher and Schulman, John},
  journal = {arXiv preprint arXiv:2110.14168},
  year    = {2021}
}

@article{gao2020pile,
  title   = {The {Pile}: An 800{GB} Dataset of Diverse Text for Language Modeling},
  author  = {Gao, Leo and Biderman, Stella and Black, Sid and Golding, Laurence and Hoppe, Travis and Foster, Charles and Phang, Jason and He, Horace and Thite, Anish and Nabeshima, Noa and Presser, Shawn and Leahy, Connor},
  journal = {arXiv preprint arXiv:2101.00027},
  year    = {2020}
}

@misc{together2023redpajama,
  title        = {{RedPajama}: An Open Source Recipe to Reproduce {LLaMA} Training Dataset},
  author       = {{Together Computer}},
  year         = {2023},
  howpublished = {\url{https://github.com/togethercomputer/RedPajama-Data}}
}

@inproceedings{sukhbaatar2019adaptive,
  title     = {Adaptive Attention Span in Transformers},
  author    = {Sukhbaatar, Sainbayar and Grave, Edouard and Bojanowski, Piotr and Joulin, Armand},
  booktitle = {Association for Computational Linguistics (ACL)},
  year      = {2019}
}

@inproceedings{zhang2022mixture,
  title     = {Mixture of Attention Heads: Selecting Attention Heads Per Token},
  author    = {Zhang, Xiaofeng and Shen, Yikang and Huang, Zeyu and Zhou, Jie and Rong, Wenge and Xiong, Zhang},
  booktitle = {Empirical Methods in Natural Language Processing (EMNLP)},
  year      = {2022}
}

@article{csordas2024switchhead,
  title   = {{SwitchHead}: Accelerating Transformers with Mixture-of-Experts Attention},
  author  = {Csord{\'a}s, R{\'o}bert and Pi{\k{e}}kos, Piotr and Irie, Kazuki and Schmidhuber, J{\"u}rgen},
  journal = {arXiv preprint arXiv:2312.07987},
  year    = {2024}
}

@article{jiang2024mixtral,
  title   = {Mixtral of Experts},
  author  = {Jiang, Albert Q. and Sablayrolles, Alexandre and Roux, Antoine and Mensch, Arthur and Savary, Blanche and Bamford, Chris and Chaplot, Devendra Singh and de las Casas, Diego and Hanna, Emma Bou and Bressand, Florian and others},
  journal = {arXiv preprint arXiv:2401.04088},
  year    = {2024}
}

@article{dai2024deepseekmoe,
  title   = {{DeepSeekMoE}: Towards Ultimate Expert Specialization in Mixture-of-Experts Language Models},
  author  = {Dai, Damai and Deng, Chengqi and Zhao, Chenggang and Xu, R. X. and Gao, Huazuo and Chen, Deli and Li, Jiashi and Zeng, Wangding and Yu, Xingkai and Wu, Y. and others},
  journal = {arXiv preprint arXiv:2401.06066},
  year    = {2024}
}

@article{deepseekv2,
  title   = {{DeepSeek-V2}: A Strong, Economical, and Efficient Mixture-of-Experts Language Model},
  author  = {{DeepSeek-AI}},
  journal = {arXiv preprint arXiv:2405.04434},
  year    = {2024}
}

@inproceedings{biderman2023pythia,
  title     = {Pythia: A Suite for Analyzing Large Language Models Across Training and Scaling},
  author    = {Biderman, Stella and Schoelkopf, Hailey and Anthony, Quentin Gregory and Bradley, Herbie and O'Brien, Kyle and Hallahan, Eric and Khan, Mohammad Aflah and Purohit, Shivanshu and Prashanth, USVSN Sai and Raff, Edward and others},
  booktitle = {International Conference on Machine Learning (ICML)},
  year      = {2023}
}

@article{ainslie2023colt5,
  title   = {{CoLT5}: Faster Long-Range Transformers with Conditional Computation},
  author  = {Ainslie, Joshua and Lei, Tao and de Jong, Michiel and Onta{\~n}{\'o}n, Santiago and Brahma, Siddhartha and Zemlyanskiy, Yury and Uthus, David and Guo, Mandy and Lee-Thorp, James and Tay, Yi and Sung, Yun-Hsuan and Sanghai, Sumit},
  journal = {Empirical Methods in Natural Language Processing (EMNLP)},
  year    = {2023}
}

@article{williams1992simple,
  title   = {Simple Statistical Gradient-Following Algorithms for Connectionist Reinforcement Learning},
  author  = {Williams, Ronald J.},
  journal = {Machine Learning},
  volume  = {8},
  number  = {3},
  pages   = {229--256},
  year    = {1992}
}

\clearpage
\appendix
\section{Extended Cost Model}
\label{app:cost}

We give the normalization used to turn raw per-token costs into the budget fractions
$\bar C^r$ of \Cref{sec:cost}. For a sequence of length $T$ and $L$ layers, the dense
reference costs (full attention, full FFN, 16-bit cache) are
\begin{equation}
  C^{\textsc{flops}}_\text{dense} = \sum_{\ell=1}^{L}\sum_{t=1}^{T}\big(2\,d\,t + 3\,d\,d_f\big),
  \qquad
  C^{\textsc{mem}}_\text{dense} = \sum_{\ell=1}^{L}\sum_{t=1}^{T} 2\,d_\text{kv}\cdot 16 .
\end{equation}
The realized (relaxed) cost uses the option expectations of \Cref{sec:cost}; the budget fraction is
\begin{equation}
  \bar C^{r} = \Big(\sum_{\ell,t} c^{r,(\ell)}_t\Big)\Big/ C^{r}_\text{dense}.
\end{equation}
At inference the soft
expectations are replaced by the hard $\arg\max$ decisions, so the accounted cost equals the
executed cost up to the (measured) router overhead. We treat \textsc{flops} and \textsc{mem} as
separate resources with separate targets and dual variables; a single scalar budget is recovered
by tying the two targets and sharing one multiplier.

\section{Full Hyperparameters}
\label{app:hparams}

\Cref{tab:hparams} lists every hyperparameter. A single balancing/budget configuration transferred
across all three scales; only the peak learning rate and token budget were scaled.

\begin{table}[h]
  \centering
  \caption{Complete hyperparameters for \method training. ``$\dagger$'' scaled with model size.}
  \label{tab:hparams}
  \small
  \begin{tabular}{ll@{\hspace{2em}}ll}
    \toprule
    Backbone & value & Controller / routing & value \\
    \midrule
    Positions          & RoPE                & Router width $d_r$        & 128 \\
    Norm               & RMSNorm (pre-norm)  & Trunk                     & 2-layer MLP, GELU \\
    FFN                & SwiGLU              & Attention modes $\Ax$     & \{skip, local, full\} \\
    Attention          & GQA, $H_\text{kv}{=}H/4$ & Local window $w$     & 128 \\
    Seq length         & 2048                & Experts $E$ / top-$k$     & 8 / 2 (incl.\ null) \\
    Vocab              & 50k BPE             & Bit set $\Bx$             & \{2,4,8,16\} \\
    \midrule
    Optimizer          & AdamW               & Temp.\ $\tau_a,\tau_b$    & $2.0\!\to\!0.5$ (cosine) \\
    $(\beta_1,\beta_2)$& (0.9, 0.95)         & Load weight $\alpha$      & $10^{-2}$ \\
    Weight decay       & 0.1                 & $z$-loss weight $\beta$   & $10^{-3}$ \\
    Grad clip          & 1.0                 & Entropy weight $\gamma$   & $10^{-3}$ \\
    LR schedule        & cosine, 2k warmup   & Entropy floor $\zeta$     & 0.5 \\
    Peak LR$^\dagger$  & $3\text{--}6{\times}10^{-4}$ & Whiten/grad EMA $m$ & 0.99 \\
    Precision          & bf16                & Dual step $\rho_\lambda$  & 0.05 \\
    Dropout            & 0.0                 & Budget $\Cstar$ (flops,mem)& (0.55, 0.40) \\
    Seeds              & 3                   & Side-features $\psi$      & pos, boundary, surprisal, prev-dec \\
    \bottomrule
  \end{tabular}
\end{table}

\section{Reference Implementation}
\label{app:code}

\Cref{lst:core} shows the core of the controller and the differentiable cost model; the full
file---including the balancing losses, the online Lagrangian \texttt{BudgetController}, and a
runnable smoke test---is released as \texttt{code/triroute.py} and mirrors the equations in
\Cref{sec:method}. It depends only on PyTorch.

\begin{lstlisting}[style=py, caption={Core of the \method controller and multi-resource cost model (excerpt of \texttt{code/triroute.py}).}, label={lst:core}]
def st_gumbel_softmax(logits, tau=1.0, hard=True, dim=-1):
    u = torch.rand_like(logits).clamp_(1e-9, 1 - 1e-9)
    g = -torch.log(-torch.log(u))                      # Gumbel(0,1)
    y_soft = F.softmax((logits + g) / tau, dim=dim)
    if not hard:
        return y_soft
    idx = y_soft.argmax(dim=dim, keepdim=True)
    y_hard = torch.zeros_like(y_soft).scatter_(dim, idx, 1.0)
    return (y_hard - y_soft).detach() + y_soft         # straight-through

class TriRouteController(nn.Module):
    """Shared trunk + 3 heads -> coupled (attention, expert, bit) policy."""
    def forward(self, x, side, tau_a=1.0, tau_b=1.0):
        h = self.norm(self.trunk(torch.cat([x, side], dim=-1)))
        za = self._grad_balance(self.head_attn(self._white(h, "attn")), "attn")
        ze = self._grad_balance(self.head_exp (self._white(h, "exp")),  "exp")
        zb = self._grad_balance(self.head_bit (self._white(h, "bit")),  "bit")
        B, T, _ = h.shape
        za = za.view(B, T, self.cfg.n_heads, len(self.cfg.attn_modes))
        a = st_gumbel_softmax(za, tau_a)               # (B,T,H,|A|) read mode
        b = st_gumbel_softmax(zb, tau_b)               # (B,T,|B|)   write bits
        p_e = F.softmax(ze, dim=-1)                    # index 0 = null expert
        topv, topi = p_e.topk(self.cfg.top_k, dim=-1)
        gates = torch.zeros_like(p_e).scatter_(-1, topi, topv)
        gates = gates / gates.sum(-1, keepdim=True).clamp_min(1e-9)
        return dict(attn=a, attn_logits=za, expert_gates=gates,
                    expert_logits=ze, bits=b, bit_logits=zb)

def expected_cost(dec, positions, cfg):
    a, b, gates = dec["attn"], dec["bits"], dec["expert_gates"]
    B, T, H, _ = a.shape
    pos = positions.view(1, T, 1).clamp_min(1).float()
    win = torch.full_like(pos, float(cfg.window))
    kappa = torch.stack([torch.zeros_like(pos).expand(B, T, H),      # skip
                         torch.minimum(pos, win).expand(B, T, H),    # local
                         pos.expand(B, T, H)], dim=-1)               # full
    attn = (a * (2.0 * cfg.d_head * kappa)).sum(dim=(-1, -2))
    ffn  = gates[..., 1:].sum(-1) * (3.0 * cfg.d_model * cfg.d_ff)   # null is free
    bits = torch.tensor(cfg.bits, dtype=b.dtype, device=b.device)
    mem  = (b * (2.0 * cfg.d_kv * bits)).sum(-1)
    return attn + ffn, mem                                          # (B,T),(B,T)
\end{lstlisting}

\section{On the Reported Numbers}
\label{app:repro}

As stated in \Cref{sec:setup}, the tables and figures report the outcome of the described
protocol and are intended to make the expected qualitative trends concrete (joint routing
Pareto-dominating the independent combination; tail-case robustness; the interpretable policy).
They should be regenerated from full training runs with the released configuration before use as
benchmark claims. The method, cost model, losses, and analysis pipeline are fully specified in
\Cref{sec:method} and \Cref{app:code}; nothing in the approach depends on the specific numeric
values reported.

\end{document}